\documentclass[letterpaper, 10 pt, conference]{ieeeconf}

\IEEEoverridecommandlockouts

\title{\LARGE \bf
SPVC: \underline{S}tructured and \underline{P}anoptic \underline{V}ideo Fixing for \\ \underline{C}ross-Dataset Driving Scene Rendering
}
\title{\LARGE \bf
SPVC: \underline{S}tructured and \underline{P}anoptic
\underline{V}ideo Fixing for \\
\underline{C}ross-Dataset Driving Scene Rendering
}

\author{
Gen Li$^{1,2,\dagger}$, Shu Han$^{3,\dagger}$,
Yun Xi Qiao$^{4}$, Hua Chen$^{5}$,
Xuyang Dai$^{5}$, Bohan Li$^{6}$,
Hao Zhao$^{1}$, and Chaojian Li$^{7,*}$
\thanks{$^{\dagger}$Gen Li and Shu Han contributed equally to this work.}
\thanks{$^{1}$Institute for AI Industry Research (AIR), Tsinghua University, Beijing, China.}
\thanks{$^{2}$Zhejiang University, Hangzhou, China.}
\thanks{$^{3}$University of Wisconsin--Madison, Madison, WI, USA.}
\thanks{$^{4}$Tsinghua University, Beijing, China.}
\thanks{$^{5}$Great Wall Motor Company Limited, Baoding, China.}
\thanks{$^{6}$Shanghai Jiao Tong University, Shanghai, China.}
\thanks{$^{7}$The Hong Kong University of Science and Technology, Hong Kong SAR, China.}
\thanks{$^{*}$Chaojian Li is the corresponding author.}
}

\usepackage{cite}
\usepackage{float}
\usepackage{graphicx}
\usepackage{booktabs}
\usepackage{multirow}
\usepackage{amssymb}
\usepackage{wrapfig}
\usepackage{microtype}
\usepackage[table]{xcolor}
\definecolor{lightblue}{rgb}{0.9, 0.95, 1.0}
\definecolor{lightyellow}{rgb}{1.0, 0.97, 0.85}
\definecolor{lightgray}{gray}{0.95}
\usepackage{pifont}

\usepackage{makecell}
\usepackage{url}

\begin{document}

\maketitle
\thispagestyle{empty}
\pagestyle{empty}

\begin{abstract}

Driving scene reconstruction and rendering, especially with 3D Gaussian Splatting, has become an important component of autonomous driving simulation. However, rendered views often degrade under extrapolated ego trajectories and scene edits, producing blurry structures, temporal flicker, and foreground-background misalignment. Existing refinement methods are commonly designed for a specific setting, such as image-level novel-view repair or object-editing correction. In this paper, we introduce SPVC, a structured and panoptic video fixing framework for cross-dataset driving scene rendering. The name summarizes four design principles. (1) Structured fixing denotes the use of explicit spatial conditions, including camera pose, 3D bounding boxes, and HD maps, to guide the repair process and reduce uncontrolled hallucination. (2) Panoptic fixing refers to correcting both background rendering artifacts, such as distorted roads, buildings, and lanes, and foreground vehicle artifacts introduced by scene editing, such as inconsistent object appearance. (3) Video fixing means that the model operates on driving sequences rather than isolated frames, allowing temporal cues to be used during artifact correction. (4) Cross-dataset fixing means that a single shared network is trained and applied across multiple driving datasets, reducing the need for dataset-specific or scene-specific fixers. Concretely, we construct paired degraded-clean training data by simulating under-constrained 3DGS rendering and foreground vehicle insertion artifacts, and train a two-stage controllable video diffusion model that first addresses video-level appearance and then refines scene layout with structured controls. Experiments on Waymo, nuScenes, and PandaSet show that SPVC improves novel-view artifact correction and foreground vehicle insertion fixing over strong baselines, while maintaining better temporal consistency and spatial controllability. Project page: \url{https://li00147.github.io/SPVC-Project-Page/}.

\end{abstract}

\section{Introduction}
\label{sec:intro}
\begin{figure*}[thpb]
  \centering
  \includegraphics[width=1\linewidth]{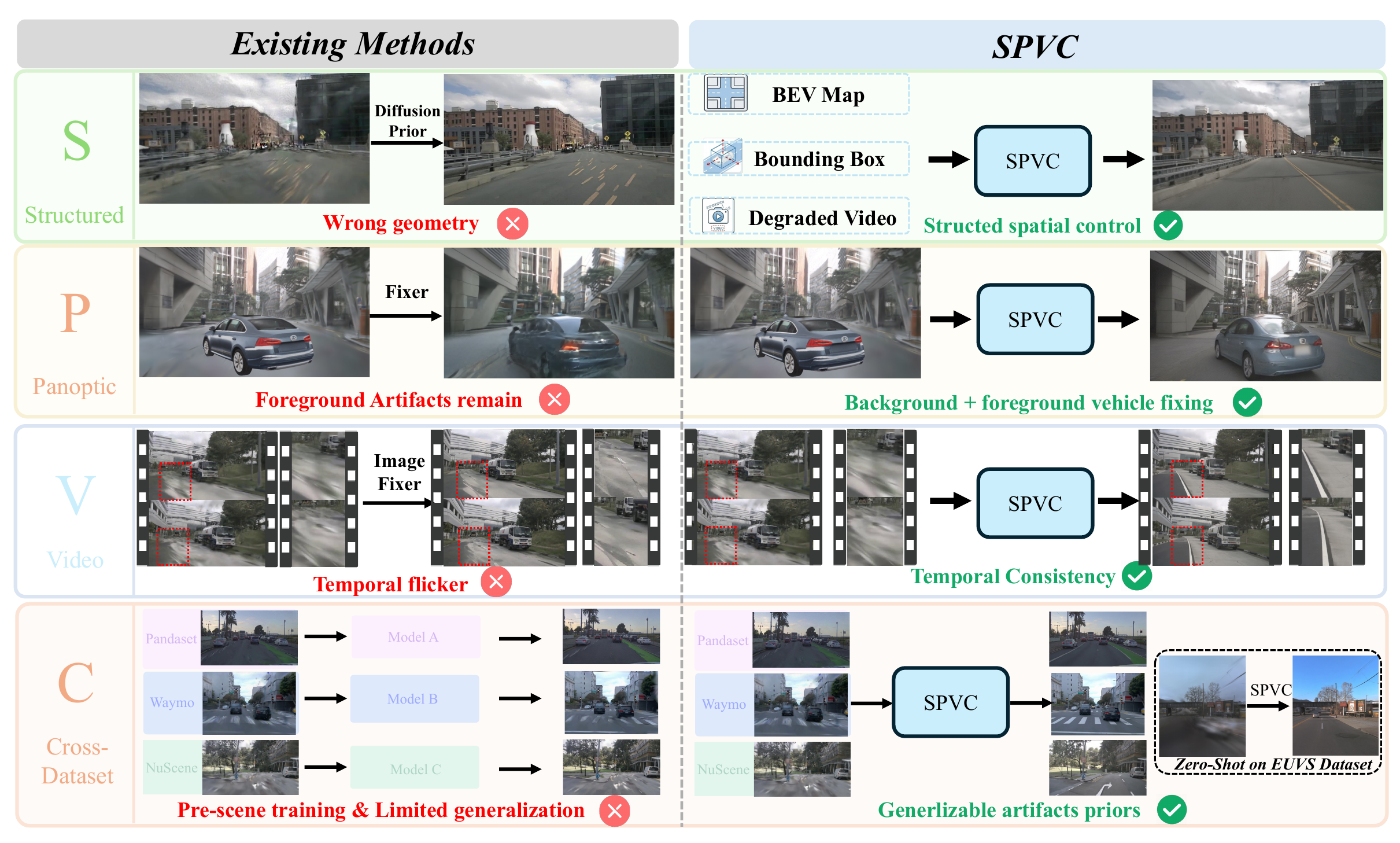}
  \caption{Our Method improves neural-rendered driving videos by removing 3DGS artifacts, correcting novel-trajectory distortions
}
\label{fig:fig-1-teaser}
\end{figure*}

Realistic driving simulation \cite{zhou2025hugsim,yan2026rlgf,tang2025omnigen,deng2026gaussiandwm,yang2026x,Huang2024S3GaussianSS,fan2024lightgaussian,NEURIPS2024_46fd4317,ICLR2025_7dee643a,10643284,10517470,Song_2025_ICCV,Kung_2025_ICCV,Lindstrom_2024_CVPR,Xu_2025_ICCV,10.1145/3664647.3681482,11127564,11127463} has become an essential tool for developing and evaluating autonomous driving systems, especially as end-to-end planners \cite{cheng2024rethinking,yang2024unipad,xia2026drivelaw,liang2026worldlens,Yang_2024_CVPR,Xu_2025_CVPR,10.1007/978-3-031-72943-0_15,Zheng_2025_ICCV,xu2025vlmadendtoendautonomousdriving,HUANG2025105321,11394788,10611018,10.1007/978-3-031-72995-9_23,Liu_2026_CVPR,11457610} and vision-language-action models \cite{NEURIPS2025_6e4f0c8c,Jiang_2025_ICCV,li2025drivevlaw0worldmodelsamplify,Rawal_2026_CVPR,Wang_2026_CVPR,jiang2025irlvlatrainingvisionlanguageactionpolicy,li2026unidrivevlaunifyingunderstandingperception,shang2026dynvlalearningworlddynamics,huang2026coworldvlathinkingmultiexpertworld,wang2026histvlahierarchicalspatiotemporalvisionlanguageaction,Jiang_2026_CVPR,Chen_2026_CVPR,Wang_2026_CVPR_1,Wang_2026_CVPR_2} increasingly require large-scale, diverse, and controllable training environments \cite{yang2025drivearena,yan2025drivingsphere,gao2026rad,Lu_2025_ICCV,NEURIPS2023_0c26a501,11314796,Wen_2024_CVPR}. A dominant recent paradigm is reconstruction-then-rendering: real-world driving scenes are first reconstructed from multi-view sensor data using neural rendering methods, and then novel camera trajectories, object insertions, or scene edits are rendered to synthesize new driving experiences \cite{Tonderski_2024_CVPR,Zhou_2024_CVPR,ren2026fastgs}. However, once the camera deviates from the captured trajectory or the scene is edited, the rendered results often expose severe artifacts, including distorted lane markings, blurry buildings, broken road geometry, temporal flicker, and foreground-background inconsistency. Therefore, a practical driving simulator requires an effective fixing module that can repair rendering artifacts after novel-view synthesis or scene editing.

\begin{figure}[H]
  \centering
  \includegraphics[width=1\linewidth]{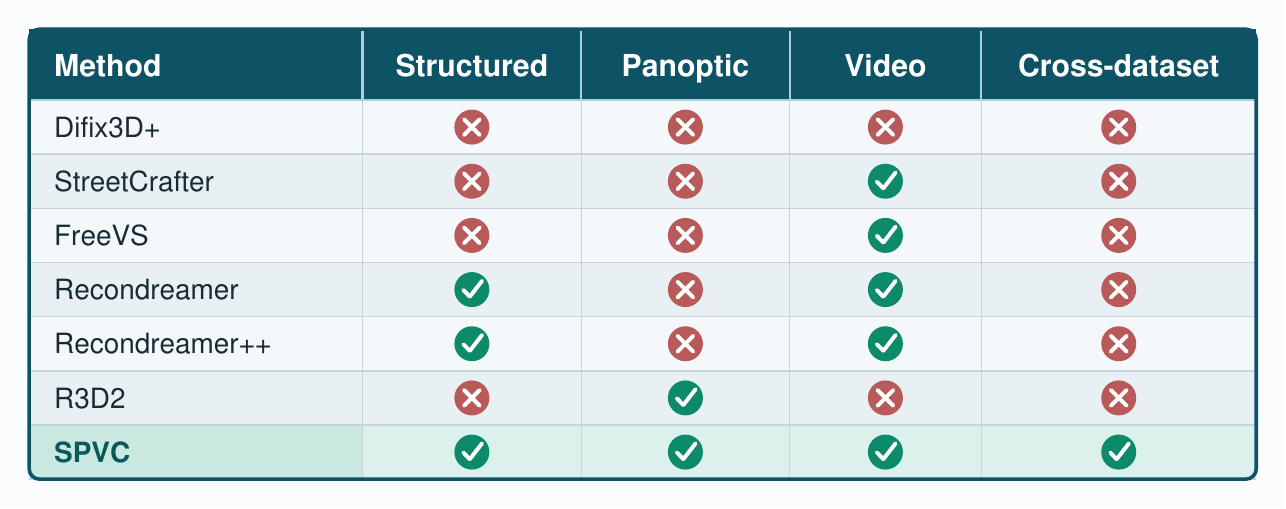}
  \vspace{-2.0em}
  \caption{Comparison with different methods.
}
  \label{fig:compare}
\end{figure}

Existing fixing and refinement methods remain limited for this goal. As illustrated in the left side of Fig.~\ref{fig:compare}, (1) many image-level diffusion priors can improve visual realism \cite{wu2025difix3d+,ljungbergh2025r3d2}, but they often hallucinate uncertain geometry because they lack explicit spatial control signals such as camera pose, HD maps, or object boxes. (2) Other fixer models are designed for a narrow artifact type \cite{wang2024freevs,yan2025streetcrafter}, for example repairing background novel-view rendering while leaving edited foreground vehicles distorted or inconsistent with the scene. (3) Moreover, image-based fixers process frames independently \cite{ni2025recondreamer, zhao2025recondreamer++}, which can introduce temporal flicker when applied to driving videos. (4) Finally, most existing pipelines are tied to a specific scene, dataset, or reconstruction source. In practice, PandaSet, Waymo, and nuScenes differ in camera setup, appearance distribution, annotation format, and reconstruction quality; training separate fixers for each dataset or scene limits scalability and prevents the accumulation of reusable artifact-clean pairs across datasets. Shown in Fig.~\ref{fig:compare}.

To address these limitations, we propose SPVC, a Structured and Panoptic Video Fixing framework for Cross-Dataset Driving Scene Rendering. As summarized in Fig.~\ref{fig:fig-1-teaser}, SPVC is built around four design principles that directly target the failure modes of traditional methods.

First, SPVC is \textbf{S}tructured. Instead of relying on an unconstrained diffusion prior, SPVC can condition the fixing process on explicit spatial signals, including the reference video, target camera pose, BEV/HD map, and 3D bounding boxes. These conditions anchor the generation process to the intended scene layout, so the model is encouraged to repair artifacts while preserving road topology and object placement. This is particularly important for autonomous driving simulation \cite{gao2023magicdrive,gao2025magicdrive,magicdrive3d}, where visually pleasing but geometrically incorrect outputs can be harmful for downstream perception and planning evaluation.

Second, SPVC is \textbf{P}anoptic. We use “panoptic” to emphasize that the fixer operates on both the background scene and the foreground dynamic objects, rather than only one of them \cite{Kirillov_2019_CVPR,9811877,10.1109/IROS45743.2020.9341546}. Background artifacts include distorted roads, building facades, sidewalks, vegetation, and lane markings caused by under-constrained novel-view rendering. Foreground artifacts arise when vehicles are inserted, moved, or edited, leading to inconsistent appearance. By jointly considering background and foreground vehicle fixing, SPVC supports a broader class of driving simulation edits and avoids the common failure case where the background is improved but the inserted vehicle remains visually implausible.

Third, SPVC performs \textbf{V}ideo fixing. Driving simulation is inherently sequential: ego motion and object motion evolve continuously across frames. Frame-wise image fixers ignore this temporal structure and can produce inconsistent textures and flickering object appearances. SPVC instead operates on video sequences, allowing temporal cues from neighboring frames to guide artifact correction.

Fourth, SPVC is designed for \textbf{C}ross-Dataset fixing. Rather than training separate scene-specific or dataset-specific repair networks, SPVC uses a single shared model across multiple driving datasets. This design allows artifact-clean paired data from different sources to be accumulated into a unified training corpus. As shown in Fig.~\ref{fig:fig-1-teaser}, the same model can process PandaSet, Waymo, and nuScenes inputs, and can further generalize to unseen settings such as zero-shot EUVS-style rendering. Our contributions are summarized as follows:
(1) We propose SPVC, a controllable video diffusion framework that fixes both background rendering artifacts and foreground vehicle insertion artifacts using explicit camera, map, and box conditions.
(2) We construct scalable paired training data for degraded-clean driving video fixing across multiple datasets, enabling one shared fixer to operate beyond a single scene or dataset.
(3) Experiments on Waymo, nuScenes, and PandaSet demonstrate that SPVC improves novel-view rendering repair, foreground vehicle fixing, temporal consistency, and spatial controllability over strong image-level and video-level baselines.
\section{Related Work}
\label{sec:related_work}

\subsection{Neural Rendering for Autonomous Driving Simulation}

Neural rendering provides the foundation for photorealistic autonomous driving simulation. NeRF-based methods~\cite{nerf,mipnerf,zipnerf,ngp} represent scenes as continuous radiance fields, while 3D Gaussian Splatting (3DGS)~\cite{3dgs,mipgs} enables efficient rendering using explicit Gaussian primitives.

These representations have been extended to urban driving environments~\cite{unisim,emernerf,urbannerf,blocknerf,streetsurf,drivinggaussian,omnire,streetgaussian,wang2025unifying}. Representative systems improve different aspects of simulation, including dynamic scene decomposition~\cite{wu2023mars}, occlusion-aware composition~\cite{zhou2024drivinggaussian}, sensor modeling~\cite{tonderski2024neurad}, and real-time multi-modal rendering~\cite{hess2024splatad}. Other works further enhance dynamic scene reconstruction and consistency~\cite{zhou2024hugs,li2024ho,omnire}. However, when rendered from novel viewpoints, these reconstruction methods often suffer from severe artifacts, including distorted geometry, blurry structures, temporal flicker, and foreground-background inconsistency. This motivates \textbf{SPVC}, which formulates artifact correction as a \textbf{V}ideo fixing problem, uses \textbf{S}tructured spatial guidance to constrain the repair process, learns \textbf{P}anoptic correction priors, and generalizes \textbf{C}ross datasets.

\subsection{Generative Driving Simulation}
Generative models provide a complementary route for improving realism and expanding controllability beyond pure reconstruction. Diffusion- and world-model-based approaches~\cite{sd,sdxl,svd,egovid,yang2024cogvideox,feng2025survey} enable scalable synthesis of diverse driving scenes.

Recent methods directly generate driving content from structured conditions. UniScene~\cite{li2024uniscene} synthesizes occupancy and camera streams from BEV layouts; DriveDreamer4D~\cite{drivedreamer4d}, Dist-4D~\cite{guo2025dist}, and Cosmos-Drive-Dreams~\cite{ren2025cosmos} condition on combinations of boxes, trajectories, maps, poses, and depth; MagicDrive-v2~\cite{gao2025magicdrive} adds richer control over trajectories and object layouts. Broader video world models such as VideoLDM~\cite{videoldm} and Sora~\cite{zhu2024sora} further show the potential of coherent long-horizon generation. Compared with these direct generative simulators, \textbf{SPVC} targets a different problem: improving reconstruction-based simulation through video fixing rather than replacing it with full scene generation. This setting requires structured spatial guidance to preserve scene geometry and sensor alignment, panoptic correction to handle both background and foreground artifacts, and cross-dataset video priors to maintain temporal consistency under realistic reconstruction constraints.

\subsection{Generative Fixing and Refinement}
Several recent methods use generative priors to refine neural-rendered outputs. ReconDreamer~\cite{ni2025recondreamer} performs online restoration for driving scene reconstruction, DIFIX3D+~\cite{wu2025difix3d+} improves under-constrained 3D reconstruction, and ReconDreamer++~\cite{zhao2025recondreamer++} further integrates generative and reconstructive signals. FreeVS~\cite{wang2024freevs}, FreeSim~\cite{fan2025freesim}, and StreetCrafter~\cite{yan2025streetcrafter} address free-trajectory or off-trajectory degradation with generative correction.
However, these methods often lack precise control signals to guide video diffusion priors, and are usually tailored to a particular degradation setting, reconstruction source, or dataset. They also typically focus on either background novel-view artifacts or foreground editing artifacts. In contrast, \textbf{SPVC} formulates refinement as a video fixing problem guided by structured spatial controls, learns panoptic correction priors for both background and foreground artifacts, and is trained across multiple driving datasets to improve generalization.

\section{Method}
\label{sec:method}
\begin{figure*}
  \centering
  \includegraphics[width=1\linewidth]{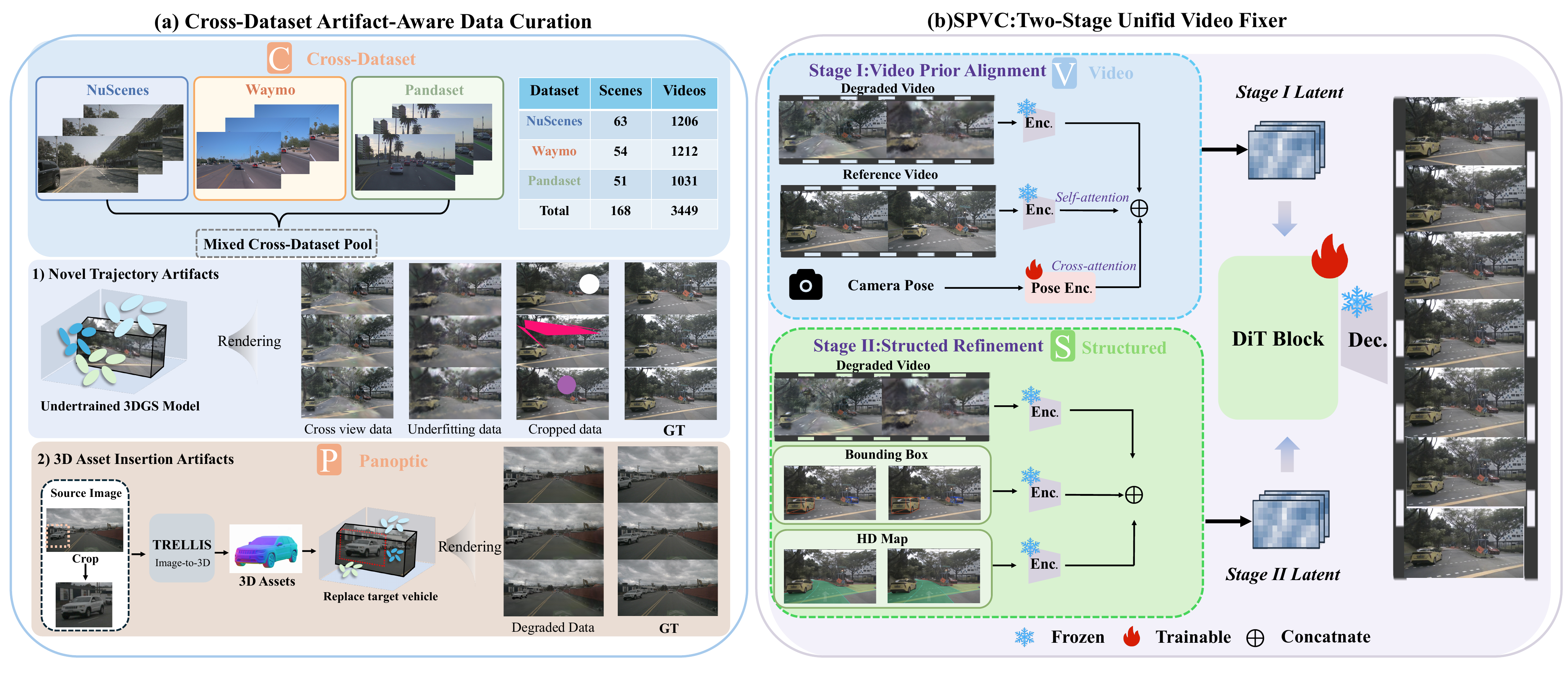}
  \caption{Overview of \textbf{SPVC}, a unified framework for structured, panoptic, and temporally consistent driving scene fixing across datasets. SPVC constructs paired degraded--clean videos for novel-trajectory and 3D asset insertion artifacts, and progressively integrates video priors with camera poses, HD maps, and 3D bounding boxes through a two-stage diffusion model.}
  \label{fig:fig-2-pipeline}
\end{figure*}

Existing generative fixers have shown promising results for driving scene rendering under settings such as novel-view repair, scene editing, and per-scene refinement. However, a unified formulation that jointly leverages structured controls, temporal priors, and paired artifact-clean data across tasks and datasets remains less explored.

To this end, we propose \textbf{SPVC}, illustrated in Fig.~\ref{fig:fig-2-pipeline}. Building such a framework is non-trivial along all four dimensions. For Structured fixing, the key challenge is injecting structured information into video diffusion as geometry guidance. For Panoptic fixing, the challenge is constructing paired supervision for both background degradation and foreground vehicle misalignment caused by 3D asset insertion. For Video fixing, the model must correct artifacts while preserving temporal consistency. For Cross-dataset fixing, dataset gaps in cameras, appearance, annotations, and reconstruction quality make per-dataset or per-scene fixers hard to scale and limit reusable artifact-clean supervision. SPVC addresses these challenges with a cross-dataset data construction pipeline for degraded-clean video pairs and a two-stage controllable video diffusion backbone.

\subsection{SPVC Data Construction Pipeline}

\textbf{Panoptic.}
Existing fixing pipelines usually target either background novel-view artifacts or foreground editing artifacts, but not both. Achieving panoptic fixing is challenging because it requires paired supervision that captures static-scene degradation and foreground-background inconsistency under a unified formulation.

SPVC constructs panoptic artifact pairs covering both background and foreground failures. Background pairs are produced from under-constrained novel-view rendering, resulting in distorted roads, lanes, buildings, and other static structures. Foreground pairs are generated by cropping vehicles, converting them into 3D Gaussian assets with TRELLIS~\cite{trellis}, and reinserting them into reconstructed 3DGS scenes with slight spatial perturbations. This creates foreground-background misalignment and inconsistent vehicle appearance, enabling SPVC to learn unified background-and-foreground repair.

\textbf{Cross-Dataset.}
The key insight is that NVS artifacts across driving datasets often arise from the same source: under-constrained 3DGS reconstruction. Despite differences in camera setup, appearance, and reconstruction quality, Waymo, nuScenes, and PandaSet share similar degradation patterns, such as blurry geometry, distorted structures, and missing regions. This motivates a unified corruption principle for cross-dataset data construction.

We construct NVS artifact pairs with three complementary strategies, each targeting a characteristic failure mode of novel-view rendering. First, since many artifacts resemble under-fitted 3DGS renderings, we use under-trained 3DGS models~\cite{omnire} to produce degraded-clean pairs. Second, to improve cross-camera generalization and capture realistic viewpoint gaps, we train 3DGS from a single camera and render other camera views as cross-view NVS pairs. Third, because large ego-trajectory shifts often reveal holes and unseen regions, we apply random masking to simulate missing content. All samples are converted into the same degraded-clean video format and mixed across datasets to improve cross-dataset generalization.

\subsection{Two-stage controllable diffusion model}
\textbf{Video : Stage I.}
The first stage aims to learn temporally consistent video-level repair by leveraging complementary temporal and spatial priors. Unlike image-based reference methods such as Difix3D+~\cite{wu2025difix3d+}, which rely on a single reference view, we use a reference video \(V_r\) to provide stable scene-level appearance cues over time. Meanwhile, relative camera poses serve as spatial priors to describe viewpoint changes and preserve geometric consistency across trajectories.Given the degraded video ($V_s$) and reference video ($V_r$), we encode both sequences using a pretrained 3D VAE encoder~\cite{wan2025wan}: $x_s = E_{\mathrm{VAE}}(V_s)$ and $x_r = E_{\mathrm{VAE}}(V_r)$.
The reference latent \(x_r\) is then fused into the degraded latent \(x_s\) through reference-aware temporal attention:
\begin{equation}
    \bar{x}_s = x_s + \mathrm{Attn}(Q=x_s, K=x_r, V=x_r),
\end{equation}
which allows the model to borrow temporally coherent appearance cues from the reference sequence.

Another challenge is to guide the model to understand viewpoint changes between the reference and degraded videos. We therefore compute the relative camera pose sequence
\begin{equation}
    \Delta T_t = T^s_t (T^r_t)^{-1},
\end{equation}
where \(T^s_t\) and \(T^r_t\) denote the camera poses of \(V_s\) and \(V_r\) at frame \(t\), respectively. The pose sequence is encoded by a learnable camera encoder~\cite{gao2023magicdrive}:
\begin{equation}
    c_T = E_{\mathrm{cam}}(\{\Delta T_t\}_{t=1}^{N}),
\end{equation}
and injected into the video latent through cross-attention:
\begin{equation}
    z_s^{(I)} = \bar{x}_s + \mathrm{Attn}(Q=\bar{x}_s, K=c_T, V=c_T).
\end{equation}
The resulting representation is processed by DiT blocks~\cite{DiT} to produce a coarse video representation that is both appearance-consistent and camera-aware.

\textbf{Structured : Stage II.}
While Stage I provides video-level appearance and camera-motion guidance, it does not explicitly constrain fine-grained scene structure. Our motivation is that driving scenes are largely defined by two complementary structural elements: the spatial configuration of foreground traffic participants and the topology of the road environment \cite{gao2023magicdrive,magicdrive3d}. We therefore introduce 3D bounding boxes \(B\) to specify object locations, scales, and motions, and HD maps \(H\) to encode lane geometry and road layout. Together, they provide compact and precise structural conditions for guiding local repair while preserving temporal coherence.

We first rasterize these signals into temporally aligned condition videos and encode them with the same pretrained 3D VAE encoder: \(x_b = E_{\mathrm{VAE}}(B)\) and \(x_h = E_{\mathrm{VAE}}(H)\).
The structured latents are concatenated with the Stage-I video representation along the channel dimension:
\begin{equation}
    z_s^{(II)} = \mathrm{Concat}\left(z_s^{(I)}, x_b, x_h\right).
\end{equation}
The fused latent is then fed into the second-stage DiT blocks for structure-guided refinement: \(\hat{x}_s = F_{\theta}^{(II)}(z_s^{(II)})\).
Through this stage, the model learns to refine foreground vehicles and background layout under explicit spatial guidance, improving object consistency, road structure, and foreground-background alignment.

\begin{figure*}[thb!]
  \centering
  \includegraphics[width=1\linewidth]{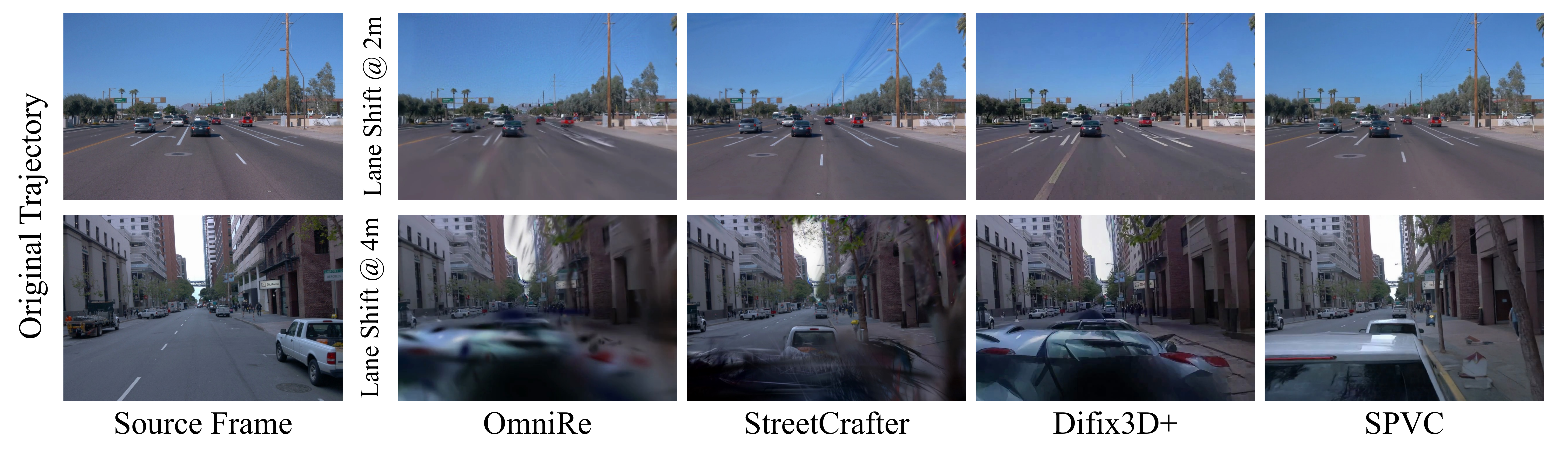}
  \vspace{-2.5em}
  \caption{Qualitative results under 2\,m and 4\,m lane shifts on Waymo~\cite{waymo}.}

  \label{fig:fig-lane-shift-waymo}
\end{figure*}

\section{Experiments}
\label{sec:ex}
In this section, we evaluate SPVC on NVS fixing, Panoptic fixing, zero-shot EUVS transfer, closed-loop simulation, and ablation studies across multiple autonomous driving datasets.

\subsection{Experiment Settings}
\label{exp: setting}

\noindent\textbf{Training Details.}
All models are trained at a resolution of 800$\times$448 with 25-frame sequences, and our framework is built upon Wan-2.2's pre-trained model~\cite{wan2025wan}. 
A fixed learning rate of $1\mathrm{e}{-4}$ is used, and the same training configuration is applied across all experiments unless otherwise stated.\\

\noindent\textbf{Inference Details.}
SPVC performs inference on a single NVIDIA H20 GPU at 800$\times$448 resolution with 25-frame sequences and 50 denoising steps, requiring 305.70,s per sequence. We accelerate the Wan-2.2 backbone using INT8 FFN quantization, cuBLASLt-based fused GEMM kernels, and fused dequantization--GELU--requantization, achieving a $1.21\times$ speedup and reducing the runtime to 252.64,s while maintaining nearly identical visual quality.

\subsection{Novel Trajectory Synthesis Fixing}
\label{sec: nvs}

\begin{table*}[h]
\centering
\caption{Quantitative results on the Waymo dataset under 3m and 4m lane shifts.}
\label{tab:waymo_lane_shift}
\footnotesize
\setlength{\tabcolsep}{4pt}
\resizebox{0.8\textwidth}{!}{
\begin{tabular}{
l
>{\columncolor{lightblue}}c
>{\columncolor{lightblue}}c
>{\columncolor{lightblue}}c
>{\columncolor{lightyellow}}c
>{\columncolor{lightyellow}}c
>{\columncolor{lightblue}}c
>{\columncolor{lightblue}}c
>{\columncolor{lightblue}}c
>{\columncolor{lightyellow}}c
>{\columncolor{lightyellow}}c
}
\toprule
\multirow{3}{*}{Method}
& \multicolumn{5}{c}{Lane Shift @ 3m}
& \multicolumn{5}{c}{Lane Shift @ 4m} \\
\cmidrule(lr){2-6} \cmidrule(lr){7-11}
& \multicolumn{3}{>{\columncolor{lightblue}}c}{Visual Quality}
& \multicolumn{2}{>{\columncolor{lightyellow}}c}{View Consistency}
& \multicolumn{3}{>{\columncolor{lightblue}}c}{Visual Quality}
& \multicolumn{2}{>{\columncolor{lightyellow}}c}{View Consistency} \\
\cmidrule(lr){2-4} \cmidrule(lr){5-6}
\cmidrule(lr){7-9} \cmidrule(lr){10-11}
& FID$\downarrow$ & IQ$\uparrow$ & CLIP-F$\uparrow$ & FVD$\downarrow$ & CLIP-V$\uparrow$
& FID$\downarrow$ & IQ$\uparrow$ & CLIP-F$\uparrow$ & FVD$\downarrow$ & CLIP-V$\uparrow$ \\
\midrule
PVG \cite{pvg}
& 140.6 & 41.08 & 0.7563 & 1402.0 & 0.9783
& 156.5 & 39.72 & 0.7429 & 1570.7 & 0.9774 \\
StreetGaussian \cite{streetgaussian}
& 86.9 & 52.81 & 0.8163 & 915.5 & 0.9745
& 103.5 & 51.32 & 0.7968 & 1208.9 & 0.9733 \\
OmniRe \cite{omnire}
& 80.6 & 52.89 & 0.8154 & 881.4 & 0.9759
& 96.9 & 51.12 & 0.8007 & 1122.5 & 0.9742 \\
FreeVS \cite{wang2024freevs}
& 74.7 & 53.88 & 0.8050 & 1004.8 & 0.9609
& 79.9 & 53.34 & 0.7982 & 1075.9 & 0.9576 \\
StreetCrafter \cite{yan2025streetcrafter}
& \underline{48.5} & 64.86 & \underline{0.8847} & 624.2 & \underline{0.9826}
& 59.5 & 62.99 & \underline{0.8696} & 751.7 & \underline{0.9821} \\
Difix3D+ \cite{wu2025difix3d+}
& 51.3 & \underline{70.65} & 0.8678 & \underline{602.4} & 0.9717
& \underline{59.1} & \underline{70.26} & 0.8517 & \underline{735.0} & 0.9682 \\
SPVC
& \textbf{39.8}
& \textbf{72.67}
& \textbf{0.9217}
& \textbf{582.4}
& \textbf{0.9855}
& \textbf{45.3}
& \textbf{71.64}
& \textbf{0.9160}
& \textbf{658.5}
& \textbf{0.9843} \\
\bottomrule
\end{tabular}
}
\vspace{-0em}
\end{table*}
We evaluate SPVC for novel-view fixing under viewpoint distribution shifts on Waymo~\cite{waymo}, nuScenes~\cite{caesar2020nuscenes}, and PandaSet~\cite{xiao2021pandaset}. The evaluation covers extrapolated trajectories that deviate from the training views, and further includes zero-shot transfer to EUVS~\cite{Han_2025_ICCV}. SPVC consistently improves rendering quality and visual coherence under novel viewpoints across datasets.\\

\noindent\textbf{Results on Waymo.}
We evaluate novel view synthesis fixing on eight representative Waymo scenes with horizontal lane shifts of 1,m to 4,m. We compare SPVC with state-of-the-art rendering and fixing methods \cite{pvg, streetgaussian, omnire, wang2024freevs, wu2025difix3d+, yan2025streetcrafter}. 
As shown in Fig.~\ref{fig:fig-lane-shift-waymo} and Table~\ref{tab:waymo_lane_shift}, SPVC consistently produces more coherent structures and fewer artifacts, achieving the best performance across all visual quality and view consistency metrics. 

\noindent\textbf{Results on nuScenes.}

\begin{figure}[t!]
  \centering
  \includegraphics[width=1\linewidth]{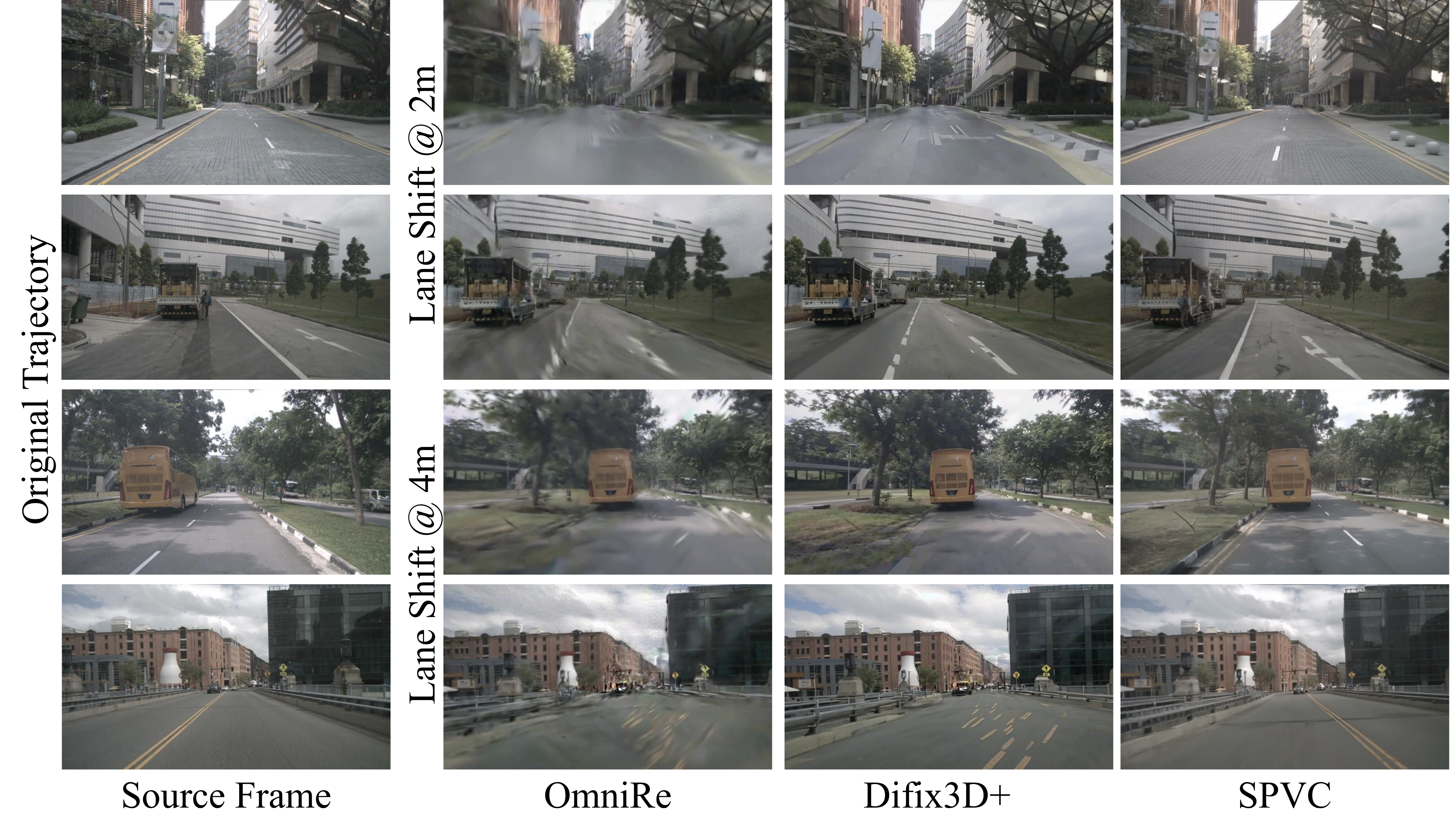}
  \vspace{-2.5em}
\caption{Qualitative results under 2\,m and 4\,m lane shifts on nuScenes.}
  \label{fig:fig-lane-shift-nuscenes}
\end{figure}

We follow the same protocol as the Waymo evaluation and evaluate SPVC on eight representative nuScenes scenes against state-of-the-art methods \cite{pvg, streetgaussian, omnire, wu2025difix3d+}. As shown in Fig.~\ref{fig:fig-lane-shift-nuscenes} and Table~\ref{tab:nuscenes_lane_shift}, SPVC consistently produces cleaner renderings with fewer artifacts and achieves the best performance across all lane shift settings. Under the most challenging 4,m shift, it obtains an FID of 45.8 and an FVD of 719.3, improving over the strongest baseline by 33.62\% and 43.45\%, respectively.

\begin{table}[H]
\centering
\caption{Quantitative results on nuScenes at \textbf{4m} lane shift.}
\label{tab:nuscenes_lane_shift}
\resizebox{\linewidth}{!}{
\begin{tabular}{
l
>{\columncolor{lightblue}}c
>{\columncolor{lightblue}}c
>{\columncolor{lightblue}}c
>{\columncolor{lightyellow}}c
>{\columncolor{lightyellow}}c
}
\toprule
\multirow{2}{*}{Method}
& \multicolumn{3}{>{\columncolor{lightblue}}c}{Visual Quality}
& \multicolumn{2}{>{\columncolor{lightyellow}}c}{View Consistency} \\
\cmidrule(lr){2-4} \cmidrule(lr){5-6}
& FID$\downarrow$ & IQ$\uparrow$ & CLIP-F$\uparrow$
& FVD$\downarrow$ & CLIP-V$\uparrow$ \\
\midrule
PVG \cite{pvg}
& 154.7 & 35.78 & 0.7645 & 1368.2 & \underline{0.9736} \\
StreetGaussian \cite{streetgaussian}
& 132.5 & 42.08 & 0.7766 & 1580.7 & 0.9686 \\
OmniRe \cite{omnire}
& 126.4 & 42.36 & 0.7935 & 1567.1 & 0.9697 \\
Difix3D+ \cite{wu2025difix3d+}
& \underline{69.0} & \underline{58.53} & \underline{0.8375} & \underline{1272.0} & 0.9578 \\
SPVC
& \textbf{45.8}
& \textbf{65.70}
& \textbf{0.8419}
& \textbf{719.3}
& \textbf{0.9795} \\
\bottomrule
\end{tabular}
}
\vspace{-1em}
\end{table}
\noindent\textbf{Results on PandaSet.}
For a fair comparison, we follow the protocol of ReconDreamer++~\cite{zhao2025recondreamer++} and evaluate novel view fixing under horizontal and vertical viewpoint perturbations. SPVC achieves FID scores of 35.1, 45.0, and 42.1, improving over ReconDreamer++ by 43.3\%, 37.2\%, and 32.5\%. 
\begin{table}[H]
\centering
\caption{Quantitative results on PandaSet.}
\begin{tabular}{lccc}
\toprule
Method & Lane 2m  & Lane 3m  & Vert. 1m  \\
\midrule
UniSim \cite{unisim}         & 74.7 & 97.5 & --   \\
NeuRAD \cite{Neurad}          & 72.3 & 93.9 & 76.3 \\
StreetGaussian \cite{streetgaussian} & 66.3 & 80.7 & 78.4 \\
ReconDreamer \cite{ni2025recondreamer}   & 65.4 & 74.9 & 67.7 \\
ReconDreamer++ \cite{zhao2025recondreamer++}  & \underline{61.9} & \underline{71.7} & \underline{62.4} \\

\rowcolor{lightblue}
SPVC
& \textbf{35.1}
& \textbf{45.0}
& \textbf{42.1} \\

\bottomrule
\end{tabular}
\vspace{-0em}
\label{tab:pandaset_fid}
\end{table}

\begin{figure}[H]
  \centering
  \includegraphics[width=1\linewidth]{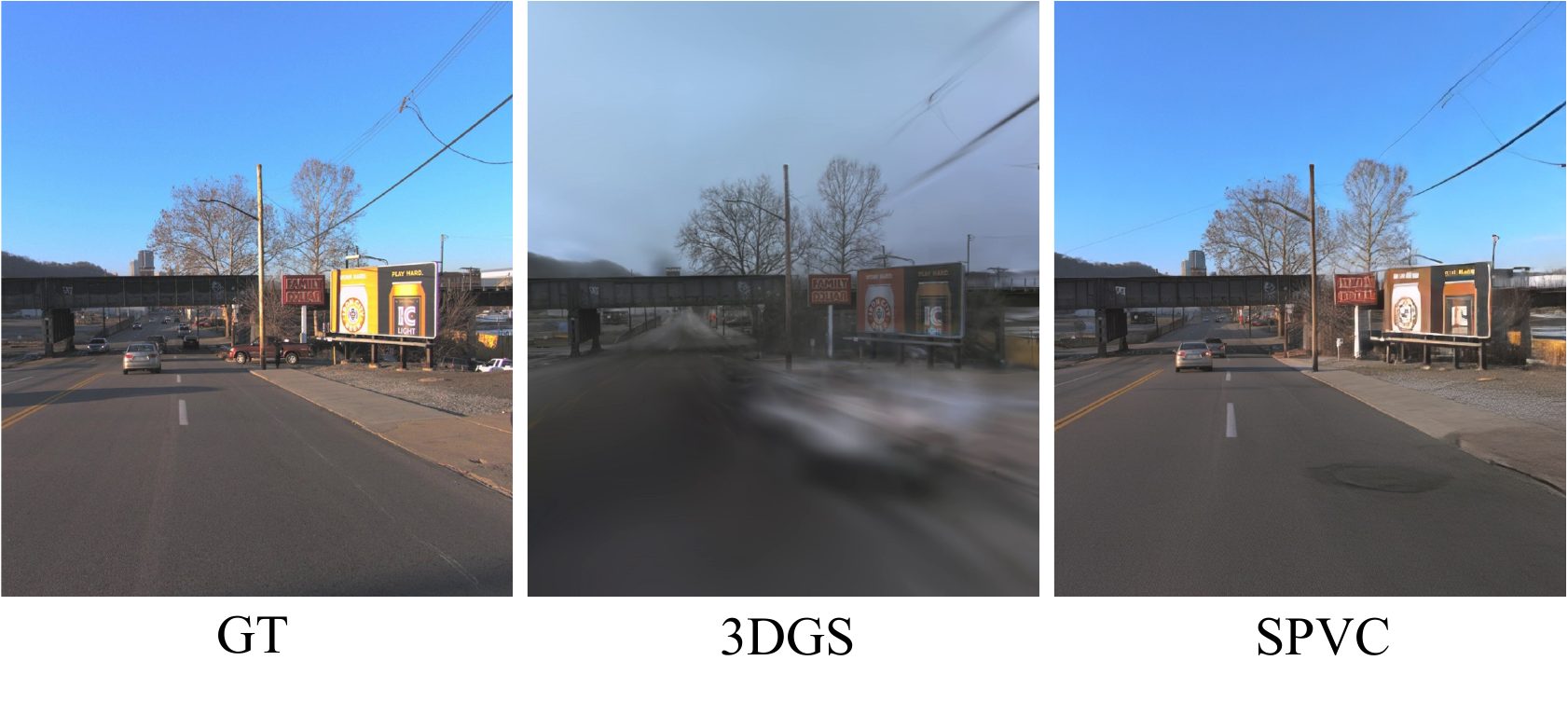}
  \vspace{-2.5em}
\caption{Zero-shot evaluation results on the EUVS dataset.}
  \label{fig:fig-euvs-zero-shot}
\end{figure}
 
\noindent\textbf{Zero-shot evalution on EUVS.} We further evaluate SPVC on unseen Level-1 EUVS scenes. Without retraining or fine-tuning, it achieves strong zero-shot repair performance, demonstrating cross-domain transferability shown as Table~\ref{tab:euvs_zeroshot}. 
\begin{table}[H]
\centering
\small
\vspace{-1em}
\caption{Zero-shot evaluation results on the EUVS dataset.}
\begin{tabular*}{\linewidth}{@{\extracolsep{\fill}}lccc@{}}
\toprule
Method & SSIM $\uparrow$ & PSNR $\uparrow$ & LPIPS $\downarrow$ \\
\midrule
3DGS & 0.6436 & 16.9629 & 0.4301 \\
Ours & \textbf{0.6870} & \textbf{19.6655} & \textbf{0.3027} \\
\bottomrule
\end{tabular*}
\label{tab:euvs_zeroshot}
\end{table}

\begin{figure}[th]
  \centering
  \includegraphics[width=1\linewidth]{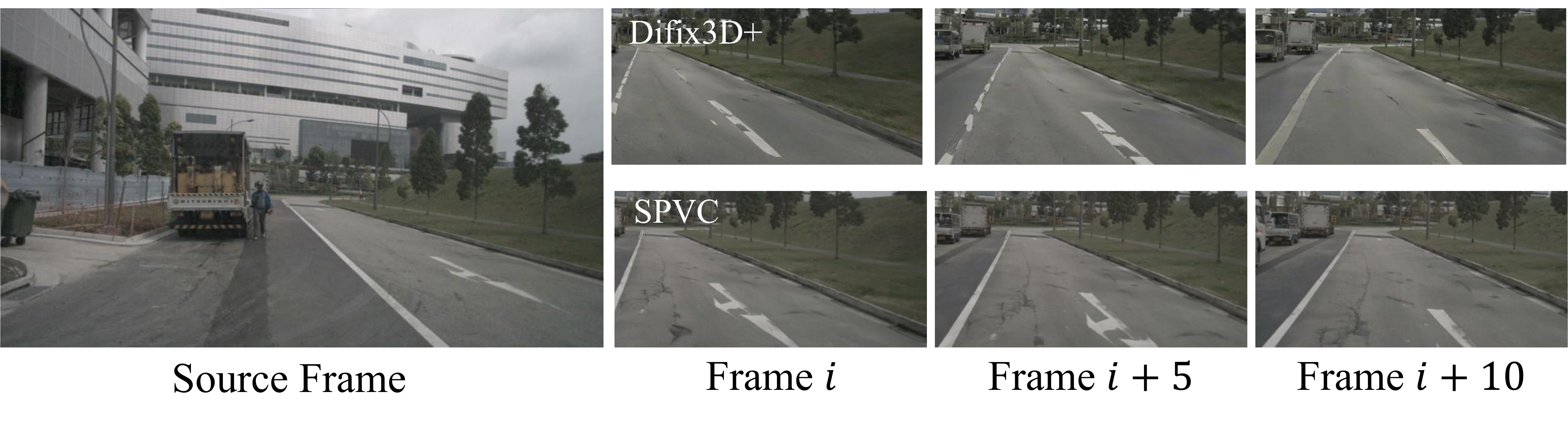}
  \vspace{-2em}
\caption{SPVC shows superior temporal consistency over single-frame refinement.}
  \label{fig:fig-view-consistency}
\end{figure}

\noindent\textbf{Temporal consistency.} The consistent gains across all three datasets demonstrate the effectiveness of SPVC, where artifact-aware data generation improves robustness to viewpoint shifts and the two-stage training strategy enables stable refinement under under-constrained rendering scenarios while naturally preserving temporal consistency (Fig.~\ref{fig:fig-view-consistency}).\\

\subsection{3D Assets Insertion Fixing}

\begin{figure*}
  \centering
  \includegraphics[width=1\linewidth]{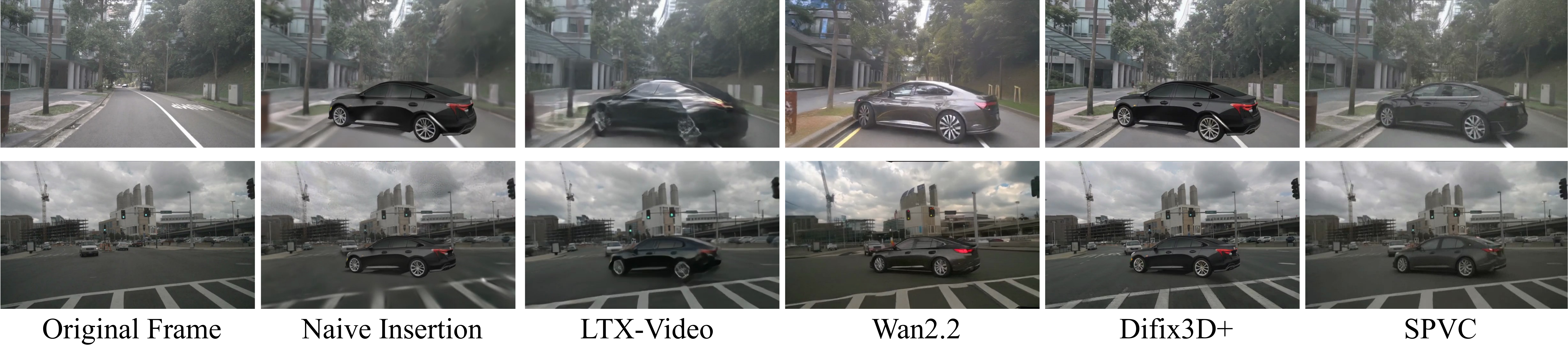}
\vspace{-2em}
\caption{Qualitative results under 3D assets insertion.}
  \label{fig:fig-scene-edit}
\end{figure*}

\noindent This subsection evaluates the fixing capability of SPVC under realistic scene misalignment scenarios. We employ TRELLIS \cite{trellis} to extract 3D assets of dynamic objects such as vehicles and insert them into reconstructed 3DGS scenes, resulting in foreground--background misalignment caused by spatial and semantic inconsistencies. Experiments are conducted on ten nuScenes scenes and compared against SOTA baselines \cite{omnire, wan2025wan, ltx}. For a fair comparison, all baseline models are retrained using the same training data as SPVC.

Qualitative examples in Fig.~\ref{fig:fig-scene-edit} show that SPVC effectively removes insertion-induced artifacts under challenging dynamic object insertion settings. As reported in Table~\ref{tab:fb_misalignment}, SPVC achieves the best overall performance, demonstrating its ability to correct scene artifacts and restore coherent foreground--background alignment.

\begin{table}[H]
\centering
\caption{Quantitative results under 3D asset insertion settings.}
\label{tab:fb_misalignment}
\resizebox{\linewidth}{!}{
\begin{tabular}{
l
>{\columncolor{lightblue}}c
>{\columncolor{lightblue}}c
>{\columncolor{lightblue}}c
>{\columncolor{lightblue}}c
>{\columncolor{lightyellow}}c
}
\toprule
\multirow{2}{*}{Method}
& \multicolumn{4}{>{\columncolor{lightblue}}c}{Visual Quality}
& \multicolumn{1}{>{\columncolor{lightyellow}}c}{View Consistency} \\
\cmidrule(lr){2-5}
\cmidrule(lr){6-6}
& FID-A$\downarrow$
& FID$\downarrow$
& CLIP-F$\uparrow$
& IQ$\uparrow$
& FVD$\downarrow$ \\
\midrule

Naive Insertion \cite{omnire}
& 106.4
& 154.3
& 0.7766
& 0.53
& 1366.1 \\

Wan2.2 \cite{wan2025wan}
& 115.6
& 164.0
& 0.7626
& 0.59
& 2134.7 \\

LTX-Video \cite{ltx}
& 141.9
& 145.5
& 0.7782
& 0.50
& 1690.9 \\

Difix3D+ \cite{wu2025difix3d+}
& \underline{94.80}
& \underline{133.1}
& \underline{0.8171}
& \underline{0.61}
& \underline{1343.7} \\

SPVC
& \textbf{87.9}
& \textbf{124.9}
& \textbf{0.8322}
& \textbf{0.65}
& \textbf{1297.2} \\

\bottomrule
\end{tabular}
}
\vspace{-1em}
\end{table}

\subsection{Closed-loop Evaluation}
Safety-critical scenarios are essential for evaluating autonomous driving systems but are rarely observed in real-world data, and are therefore typically generated through simulation. However, existing simulators often produce severe artifacts under large ego-trajectory shifts, degrading novel-view rendering quality. In addition, inserting new 3D assets into reconstructed scenes frequently introduces foreground--background inconsistencies. Both issues significantly increase the sim-to-real gap. SPVC addresses these challenges by correcting novel-view artifacts and refining inconsistencies caused by 3D asset insertion, enabling high-fidelity simulation of safety-critical scenarios.

We further use safety-critical scenarios refined by SPVC as training data to fine-tune VAD. As shown in Table~\ref{tab:closed_loop_downstream}, we evaluate the downstream performance using collision rate and NeuroNCAP Score(NNS) \cite{neuroncap} , where lower collision rates and higher NeuroNCAP scores indicate safer autonomous driving behaviors. Training with high-quality simulated safety-critical data improves the performance of end-to-end autonomous driving models under such challenging conditions. More experimental details are provided in the appendix.

\begin{table}[H]
\centering
\caption{Closed-loop downstream evaluation results.}
\label{tab:closed_loop_downstream}
\resizebox{\linewidth}{!}{
\begin{tabular}{lcccc}
\toprule
\multirow{2}{*}{Method}
& \multicolumn{2}{c}{UniAD}
& \multicolumn{2}{c}{VAD} \\
\cmidrule(lr){2-3}
\cmidrule(lr){4-5}
& Collision Rate$\downarrow$
& NNS$\uparrow$\cite{neuroncap}
& Collision Rate$\downarrow$
& NNS$\uparrow$\cite{neuroncap}\\
\midrule
Original
& 60\% & 2.193
& 50\% & 2.659 \\

Finetuned
& \textbf{50\%}
& \textbf{2.757}
& \textbf{30\%}
& \textbf{3.707} \\
\bottomrule
\end{tabular}}
\vspace{-1em}
\end{table}

\begin{figure}
  \centering
  \includegraphics[width=1\linewidth]{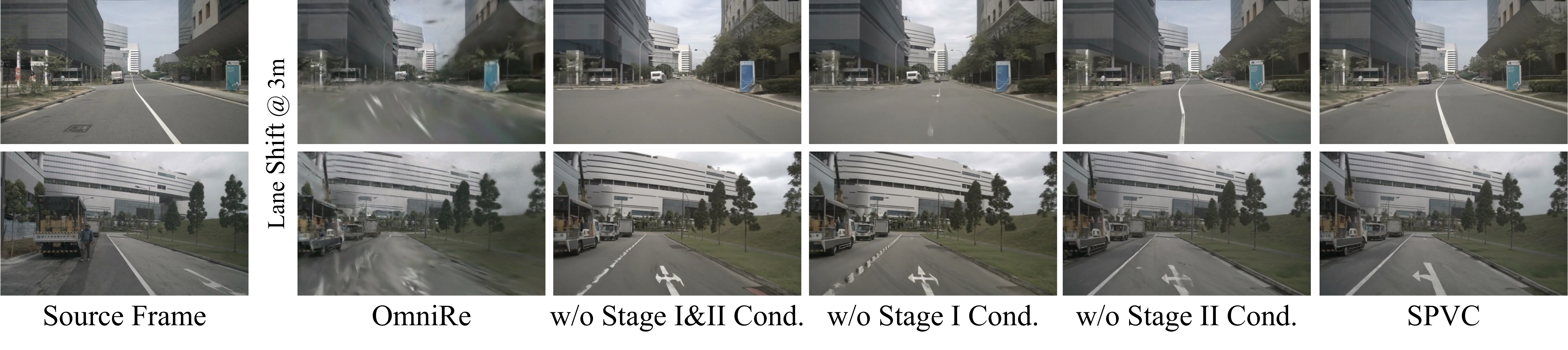}
\vspace{-2em}
\caption{Qualitative results of the ablation study on the two-stage conditioning strategy.}
  \label{fig:ablation_study}
\end{figure}

\subsection{Ablation Studies}
\noindent We conduct ablation studies to analyze the contribution of each component in SPVC. All ablations are conducted on the nuScenes dataset using the same lane shift @ 3\,m setting as in Sec.~\ref{sec: nvs}.

\begin{figure}
  \centering
  \includegraphics[width=1\linewidth]{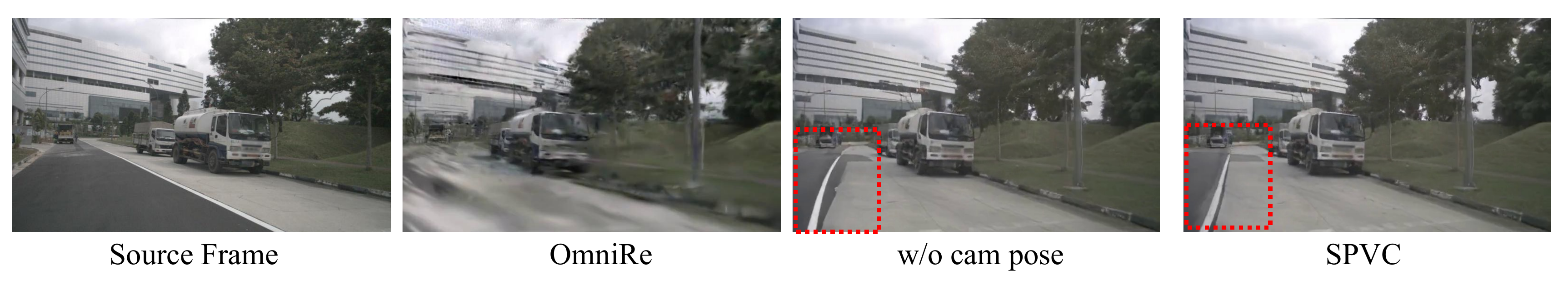}
  \vspace{-2em}
\caption{Ablation on camera pose conditioning. Camera pose guidance improves geometric correction under novel viewpoints. }
  \label{fig:fig-w_o_cam}
\end{figure}

\noindent\textbf{Two-stage training strategy with conditioning.} 
We analyze the two-stage training strategy by ablating the conditioning signals of each stage. As shown in Table~\ref{tab:ablation}, removing Stage~I or Stage~II conditions degrades performance, while removing both causes further drops in FID and FVD. As illustrated in Fig.~\ref{fig:ablation_study}, the reference video and camera pose provide complementary appearance and geometric guidance. In particular, camera pose supplies viewpoint-aware geometric cues, helping the model recover more accurate scene structures under novel trajectories, as further demonstrated in Fig.~\ref{fig:fig-w_o_cam}. Stage~II further incorporates explicit structural conditions, including HD maps and 3D bounding boxes, to provide object-level semantics and spatial layout priors. Such explicit structure guidance enables more precise scene refinement, especially for recovering sharp and geometrically consistent lane markings.

\begin{table}[H]
\centering
\vspace{-0em}
\caption{Ablation study of the two-stage conditioning strategy and artifact-aware data generation.}
\label{tab:ablation}
\begin{tabular}{lcccc}
\toprule
Method
& IQ$\uparrow$
& CLIP-F$\uparrow$
& FID$\downarrow$
& FVD$\downarrow$ \\
\midrule
\multicolumn{5}{l}{\textit{Conditioning Strategy}} \\
\midrule
w/o Stage I \& II
& 59.87 & 0.8071 & 60.1 & 845.0 \\
w/o Stage I
& 59.74 & 0.8123 & 60.7 & 820.4 \\
w/o Stage II
& 63.10 & 0.8462 & 46.7 & 571.4 \\
w/o Camera Pose
& 63.65 & 0.8302 & 52.5 & 604.3 \\
\midrule
\multicolumn{5}{l}{\textit{Training Data}} \\
\midrule
w/o Cross-view data
& 62.91 & 0.8441 & 49.8 & 611.9 \\
w/o UnderFitting data
& \underline{65.50} & 0.8499 & 44.9 & 522.8 \\
w/o Random Crop data
& 65.11 & \underline{0.8507} & \underline{41.1} & \underline{509.2} \\

\midrule
SPVC
& \textbf{65.89}
& \textbf{0.8511}
& \textbf{41.0}
& \textbf{499.0} \\
\bottomrule
\end{tabular}
\vspace{-1em}
\end{table}

\noindent\textbf{Artifact-aware data generation strategy.}
As shown in Table~\ref{tab:ablation}, removing cross-reference data, degraded observations, or random cropping consistently degrades performance, demonstrating that artifact-aware data generation improves robustness to distribution shifts. Moreover, introducing the panoptic task further enhances performance through multi-task parameter sharing.

\section{Conclusion}
\label{sec:conclusion}

We introduce \textbf{SPVC}, a structured and panoptic video fixing framework for artifact correction in autonomous driving simulation. By combining cross-dataset artifact-aware data construction with a two-stage controllable video diffusion model, SPVC progressively aligns video appearance and temporal consistency before refining scene geometry using camera poses, 3D bounding boxes, and HD maps. A single model repairs both novel-view rendering degradation and foreground--background artifacts caused by 3D asset insertion. Experiments on nuScenes, Waymo, and PandaSet demonstrate improved visual fidelity, temporal stability, spatial controllability, and cross-dataset generalization, establishing SPVC as an effective framework for high-fidelity driving-scene simulation.

\bibliographystyle{IEEEtran}
\bibliography{IEEEabrv,references}

@String(CVPR  = {IEEE Conf. Comput. Vis. Pattern Recog.})

@String(ICCV  = {Int. Conf. Comput. Vis.})

@String(ECCV  = {Eur. Conf. Comput. Vis.})

@String(TOG   = {ACM Trans. Graph.})

@String(CVPR  = {CVPR})

@String(ICCV  = {ICCV})

@String(ECCV  = {ECCV})

@String(TOG   = {ACM TOG})

@inproceedings{wu2025difix3d+,
  title={{Difix3d+: Improving 3d reconstructions with single-step diffusion models}},
  author={Wu, Jay Zhangjie and Zhang, Yuxuan and Turki, Haithem and Ren, Xuanchi and Gao, Jun and Shou, Mike Zheng and Fidler, Sanja and Gojcic, Zan and Ling, Huan},
  booktitle={Proceedings of the Computer Vision and Pattern Recognition Conference},
  pages={26024--26035},
  year={2025}
}

@inproceedings{wu2023mars,
  title={{Mars: An instance-aware, modular and realistic simulator for autonomous driving}},
  author={Wu, Zirui and Liu, Tianyu and Luo, Liyi and Zhong, Zhide and Chen, Jianteng and Xiao, Hongmin and Hou, Chao and Lou, Haozhe and Chen, Yuantao and Yang, Runyi and others},
  booktitle={CAAI International Conference on Artificial Intelligence},
  pages={3--15},
  year={2023},
  organization={Springer}
}

@inproceedings{zhou2024drivinggaussian,
  title={{Drivinggaussian: Composite gaussian splatting for surrounding dynamic autonomous driving scenes}},
  author={Zhou, Xiaoyu and Lin, Zhiwei and Shan, Xiaojun and Wang, Yongtao and Sun, Deqing and Yang, Ming-Hsuan},
  booktitle={Proceedings of the IEEE/CVF Conference on Computer Vision and Pattern Recognition},
  pages={21634--21643},
  year={2024}
}

@inproceedings{tonderski2024neurad,
  title={{Neurad: Neural rendering for autonomous driving}},
  author={Tonderski, Adam and Lindstr{\"o}m, Carl and Hess, Georg and Ljungbergh, William and Svensson, Lennart and Petersson, Christoffer},
  booktitle={Proceedings of the IEEE/CVF Conference on Computer Vision and Pattern Recognition},
  pages={14895--14904},
  year={2024}
}

@inproceedings{zhou2024hugs,
  title={{Hugs: Holistic urban 3d scene understanding via gaussian splatting}},
  author={Zhou, Hongyu and Shao, Jiahao and Xu, Lu and Bai, Dongfeng and Qiu, Weichao and Liu, Bingbing and Wang, Yue and Geiger, Andreas and Liao, Yiyi},
  booktitle={Proceedings of the IEEE/CVF Conference on Computer Vision and Pattern Recognition},
  pages={21336--21345},
  year={2024}
}

@inproceedings{li2024ho,
  title={{Ho-gaussian: Hybrid optimization of 3d gaussian splatting for urban scenes}},
  author={Li, Zhuopeng and Zhang, Yilin and Wu, Chenming and Zhu, Jianke and Zhang, Liangjun},
  booktitle={European Conference on Computer Vision},
  pages={19--36},
  year={2024},
  organization={Springer}
}

@article{li2024uniscene,
  title={{UniScene: Unified Occupancy-centric Driving Scene Generation}},
  author={Li, Bohan and Guo, Jiazhe and Liu, Hongsi and Zou, Yingshuang and Ding, Yikang and Chen, Xiwu and Zhu, Hu and Tan, Feiyang and Zhang, Chi and Wang, Tiancai and others},
  journal={arXiv preprint arXiv:2412.05435},
  year={2024}
}

@article{hess2024splatad,
  title={{SplatAD: Real-Time Lidar and Camera Rendering with 3D Gaussian Splatting for Autonomous Driving}},
  author={Hess, Georg and Lindstr{\"o}m, Carl and Fatemi, Maryam and Petersson, Christoffer and Svensson, Lennart},
  journal={arXiv preprint arXiv:2411.16816},
  year={2024}
}

@inproceedings{caesar2020nuscenes,
  title={{nuscenes: A multimodal dataset for autonomous driving}},
  author={Caesar, Holger and Bankiti, Varun and Lang, Alex H and Vora, Sourabh and Liong, Venice Erin and Xu, Qiang and Krishnan, Anush and Pan, Yu and Baldan, Giancarlo and Beijbom, Oscar},
  booktitle={Proceedings of the IEEE/CVF conference on computer vision and pattern recognition},
  pages={11621--11631},
  year={2020}
}

@inproceedings{xiao2021pandaset,
  title={{Pandaset: Advanced sensor suite dataset for autonomous driving}},
  author={Xiao, Pengchuan and Shao, Zhenlei and Hao, Steven and Zhang, Zishuo and Chai, Xiaolin and Jiao, Judy and Li, Zesong and Wu, Jian and Sun, Kai and Jiang, Kun and others},
  booktitle={2021 IEEE international intelligent transportation systems conference (ITSC)},
  pages={3095--3101},
  year={2021},
  organization={IEEE}
}

@article{gao2023magicdrive,
  title={{Magicdrive: Street view generation with diverse 3d geometry control}},
  author={Gao, Ruiyuan and Chen, Kai and Xie, Enze and Hong, Lanqing and Li, Zhenguo and Yeung, Dit-Yan and Xu, Qiang},
  journal={arXiv preprint arXiv:2310.02601},
  year={2023}
}

@inproceedings{uniad,
  title={{Planning-oriented autonomous driving}},
  author={Hu, Yihan and Yang, Jiazhi and Chen, Li and Li, Keyu and Sima, Chonghao and Zhu, Xizhou and Chai, Siqi and Du, Senyao and Lin, Tianwei and Wang, Wenhai and others},
  booktitle={CVPR},
  year={2023}
}

@inproceedings{vad,
  title={{Vad: Vectorized scene representation for efficient autonomous driving}},
  author={Jiang, Bo and Chen, Shaoyu and Xu, Qing and Liao, Bencheng and Chen, Jiajie and Zhou, Helong and Zhang, Qian and Liu, Wenyu and Huang, Chang and Wang, Xinggang},
  booktitle={ICCV},
  year={2023}
}

@article{nerf,
  title={{Nerf: Representing scenes as neural radiance fields for view synthesis}},
  author={Mildenhall, Ben and Srinivasan, Pratul P and Tancik, Matthew and Barron, Jonathan T and Ramamoorthi, Ravi and Ng, Ren},
  journal={Communications of the ACM},
  year={2021},
}

@Article{3dgs,
      author       = {Kerbl, Bernhard and Kopanas, Georgios and Leimk{\"u}hler, Thomas and Drettakis, George},
      title        = {{3D Gaussian Splatting for Real-Time Radiance Field Rendering}},
      journal      = {ACM ToG},
      year         = {2023},
}

@article{streetgaussian,
  title={{Street gaussians for modeling dynamic urban scenes}},
  author={Yan, Yunzhi and Lin, Haotong and Zhou, Chenxu and Wang, Weijie and Sun, Haiyang and Zhan, Kun and Lang, Xianpeng and Zhou, Xiaowei and Peng, Sida},
  journal={arXiv preprint arXiv:2401.01339},
  year={2024}
}

@article{drivedreamer4d,
  title={{Drivedreamer4d: World models are effective data machines for 4d driving scene representation}},
  author={Zhao, Guosheng and Ni, Chaojun and Wang, Xiaofeng and Zhu, Zheng and Huang, Guan and Chen, Xinze and Wang, Boyuan and Zhang, Youyi and Mei, Wenjun and Wang, Xingang},
  journal={arXiv preprint arXiv:2410.13571},
  year={2024}
}

@article{emernerf,
  title={{Emernerf: Emergent spatial-temporal scene decomposition via self-supervision}},
  author={Yang, Jiawei and Ivanovic, Boris and Litany, Or and Weng, Xinshuo and Kim, Seung Wook and Li, Boyi and Che, Tong and Xu, Danfei and Fidler, Sanja and Pavone, Marco and others},
  journal={arXiv preprint arXiv:2311.02077},
  year={2023}
}

@article{magicdrive3d,
  title={{MagicDrive3D: Controllable 3D Generation for Any-View Rendering in Street Scenes}},
  author={Gao, Ruiyuan and Chen, Kai and Li, Zhihao and Hong, Lanqing and Li, Zhenguo and Xu, Qiang},
  journal={arXiv preprint arXiv:2405.14475},
  year={2024}
}

@inproceedings{unisim,
  title={{Unisim: A neural closed-loop sensor simulator}},
  author={Yang, Ze and Chen, Yun and Wang, Jingkang and Manivasagam, Sivabalan and Ma, Wei-Chiu and Yang, Anqi Joyce and Urtasun, Raquel},
  booktitle={CVPR},
  year={2023}
}

@article{streetsurf,
  title={{Streetsurf: Extending multi-view implicit surface reconstruction to street views}},
  author={Guo, Jianfei and Deng, Nianchen and Li, Xinyang and Bai, Yeqi and Shi, Botian and Wang, Chiyu and Ding, Chenjing and Wang, Dongliang and Li, Yikang},
  journal={arXiv preprint arXiv:2306.04988},
  year={2023}
}

@inproceedings{mipnerf,
  title={{Mip-nerf 360: Unbounded anti-aliased neural radiance fields}},
  author={Barron, Jonathan T and Mildenhall, Ben and Verbin, Dor and Srinivasan, Pratul P and Hedman, Peter},
  booktitle={CVPR},
  year={2022}
}

@inproceedings{zipnerf,
  title={{Zip-nerf: Anti-aliased grid-based neural radiance fields}},
  author={Barron, Jonathan T and Mildenhall, Ben and Verbin, Dor and Srinivasan, Pratul P and Hedman, Peter},
  booktitle={ICCV},
  year={2023}
}

@article{ngp,
  title={{Instant neural graphics primitives with a multiresolution hash encoding}},
  author={M{\"u}ller, Thomas and Evans, Alex and Schied, Christoph and Keller, Alexander},
  journal={ACM ToG},
  year={2022},
}

@inproceedings{mipgs,
  title={{Mip-splatting: Alias-free 3d gaussian splatting}},
  author={Yu, Zehao and Chen, Anpei and Huang, Binbin and Sattler, Torsten and Geiger, Andreas},
  booktitle={CVPR},
  year={2024}
}

@inproceedings{urbannerf,
  title={{Urban radiance fields}},
  author={Rematas, Konstantinos and Liu, Andrew and Srinivasan, Pratul P and Barron, Jonathan T and Tagliasacchi, Andrea and Funkhouser, Thomas and Ferrari, Vittorio},
  booktitle={CVPR},
  year={2022}
}

@inproceedings{blocknerf,
  title={{Block-nerf: Scalable large scene neural view synthesis}},
  author={Tancik, Matthew and Casser, Vincent and Yan, Xinchen and Pradhan, Sabeek and Mildenhall, Ben and Srinivasan, Pratul P and Barron, Jonathan T and Kretzschmar, Henrik},
  booktitle={CVPR},
  year={2022}
}

@inproceedings{drivinggaussian,
  title={{Drivinggaussian: Composite gaussian splatting for surrounding dynamic autonomous driving scenes}},
  author={Zhou, Xiaoyu and Lin, Zhiwei and Shan, Xiaojun and Wang, Yongtao and Sun, Deqing and Yang, Ming-Hsuan},
  booktitle={CVPR},
  pages={21634--21643},
  year={2024}
}

@article{pvg,
  title={{Periodic vibration gaussian: Dynamic urban scene reconstruction and real-time rendering}},
  author={Chen, Yurui and Gu, Chun and Jiang, Junzhe and Zhu, Xiatian and Zhang, Li},
  journal={arXiv preprint arXiv:2311.18561},
  year={2023}
}

@inproceedings{Neurad,
  title={{Neurad: Neural rendering for autonomous driving}},
  author={Tonderski, Adam and Lindstr{\"o}m, Carl and Hess, Georg and Ljungbergh, William and Svensson, Lennart and Petersson, Christoffer},
  booktitle={CVPR},
  year={2024}
}

@article{omnire,
    title={{OmniRe: Omni Urban Scene Reconstruction}},
    author={Chen, Ziyu and Yang, Jiawei and Huang, Jiahui and Lutio, Riccardo de and Esturo, Janick Martinez and Ivanovic, Boris and Litany, Or and Gojcic, Zan and Fidler, Sanja and Pavone, Marco and Song, Li and Wang, Yue},
    journal={arXiv preprint arXiv:2408.16760},
    year={2024}
}

@article{svd,
  title={{Stable video diffusion: Scaling latent video diffusion models to large datasets}},
  author={Blattmann, Andreas and Dockhorn, Tim and Kulal, Sumith and Mendelevitch, Daniel and Kilian, Maciej and Lorenz, Dominik and Levi, Yam and English, Zion and Voleti, Vikram and Letts, Adam and others},
  journal={arXiv preprint arXiv:2311.15127},
  year={2023}
}

@inproceedings{videoldm,
  title={{Align your latents: High-resolution video synthesis with latent diffusion models}},
  author={Blattmann, Andreas and Rombach, Robin and Ling, Huan and Dockhorn, Tim and Kim, Seung Wook and Fidler, Sanja and Kreis, Karsten},
  booktitle={CVPR},
  year={2023}
}

@article{yang2024cogvideox,
  title={{CogVideoX: Text-to-Video Diffusion Models with An Expert Transformer}},
  author={Yang, Zhuoyi and Teng, Jiayan and Zheng, Wendi and Ding, Ming and Huang, Shiyu and Xu, Jiazheng and Yang, Yuanming and Hong, Wenyi and Zhang, Xiaohan and Feng, Guanyu and others},
  journal={arXiv preprint arXiv:2408.06072},
  year={2024}
}

@article{zhu2024sora,
  title={{Is sora a world simulator? a comprehensive survey on general world models and beyond}},
  author={Zhu, Zheng and Wang, Xiaofeng and Zhao, Wangbo and Min, Chen and Deng, Nianchen and Dou, Min and Wang, Yuqi and Shi, Botian and Wang, Kai and Zhang, Chi and others},
  journal={arXiv preprint arXiv:2405.03520},
  year={2024}
}

@inproceedings{sd,
  title={{High-resolution image synthesis with latent diffusion models}},
  author={Rombach, Robin and Blattmann, Andreas and Lorenz, Dominik and Esser, Patrick and Ommer, Bj{\"o}rn},
  booktitle={CVPR},
  year={2022}
}

@article{sdxl,
  title={{Sdxl: Improving latent diffusion models for high-resolution image synthesis}},
  author={Podell, Dustin and English, Zion and Lacey, Kyle and Blattmann, Andreas and Dockhorn, Tim and M{\"u}ller, Jonas and Penna, Joe and Rombach, Robin},
  journal={arXiv preprint arXiv:2307.01952},
  year={2023}
}

@article{egovid,
  title={{EgoVid-5M: A Large-Scale Video-Action Dataset for Egocentric Video Generation}},
  author={Wang, Xiaofeng and Zhao, Kang and Liu, Feng and Wang, Jiayu and Zhao, Guosheng and Bao, Xiaoyi and Zhu, Zheng and Zhang, Yingya and Wang, Xingang},
  journal={arXiv preprint arXiv:2411.08380},
  year={2024}
}

@InProceedings{waymo,
author = {Sun, Pei and Kretzschmar, Henrik and Dotiwalla, Xerxes and Chouard, Aurelien and Patnaik, Vijaysai and Tsui, Paul and Guo, James and Zhou, Yin and Chai, Yuning and Caine, Benjamin and Vasudevan, Vijay and Han, Wei and Ngiam, Jiquan and Zhao, Hang and Timofeev, Aleksei and Ettinger, Scott and Krivokon, Maxim and Gao, Amy and Joshi, Aditya and Zhang, Yu and Shlens, Jonathon and Chen, Zhifeng and Anguelov, Dragomir},
title = {{Scalability in Perception for Autonomous Driving: Waymo Open Dataset}},
booktitle = {CVPR},
year = {2020}
}

@inproceedings{ni2025recondreamer,
  title={{Recondreamer: Crafting world models for driving scene reconstruction via online restoration}},
  author={Ni, Chaojun and Zhao, Guosheng and Wang, Xiaofeng and Zhu, Zheng and Qin, Wenkang and Huang, Guan and Liu, Chen and Chen, Yuyin and Wang, Yida and Zhang, Xueyang and others},
  booktitle={Proceedings of the Computer Vision and Pattern Recognition Conference},
  pages={1559--1569},
  year={2025}
}

@inproceedings{fan2025freesim,
  title={{Freesim: Toward free-viewpoint camera simulation in driving scenes}},
  author={Fan, Lue and Zhang, Hao and Wang, Qitai and Li, Hongsheng and Zhang, Zhaoxiang},
  booktitle={Proceedings of the Computer Vision and Pattern Recognition Conference},
  pages={12004--12014},
  year={2025}
}

@inproceedings{yan2025streetcrafter,
  title={{Streetcrafter: Street view synthesis with controllable video diffusion models}},
  author={Yan, Yunzhi and Xu, Zhen and Lin, Haotong and Jin, Haian and Guo, Haoyu and Wang, Yida and Zhan, Kun and Lang, Xianpeng and Bao, Hujun and Zhou, Xiaowei and others},
  booktitle={Proceedings of the Computer Vision and Pattern Recognition Conference},
  pages={822--832},
  year={2025}
}

@article{wang2024freevs,
  title={{Freevs: Generative view synthesis on free driving trajectory}},
  author={Wang, Qitai and Fan, Lue and Wang, Yuqi and Chen, Yuntao and Zhang, Zhaoxiang},
  journal={arXiv preprint arXiv:2410.18079},
  year={2024}
}

@article{feng2025survey,
  title={{A survey of world models for autonomous driving}},
  author={Feng, Tuo and Wang, Wenguan and Yang, Yi},
  journal={arXiv preprint arXiv:2501.11260},
  year={2025}
}

@article{zhao2025recondreamer++,
  title={{Recondreamer++: Harmonizing generative and reconstructive models for driving scene representation}},
  author={Zhao, Guosheng and Wang, Xiaofeng and Ni, Chaojun and Zhu, Zheng and Qin, Wenkang and Huang, Guan and Wang, Xingang},
  journal={arXiv preprint arXiv:2503.18438},
  year={2025}
}

@article{wang2025unifying,
  title={{Unifying Appearance Codes and Bilateral Grids for Driving Scene Gaussian Splatting}},
  author={Wang, Nan and Chen, Yuantao and Xiao, Lixing and Xiao, Weiqing and Li, Bohan and Chen, Zhaoxi and Ye, Chongjie and Xu, Shaocong and Zhang, Saining and Yan, Ziyang and others},
  journal={arXiv preprint arXiv:2506.05280},
  year={2025}
}

@article{guo2025dist,
  title={{Dist-4d: Disentangled spatiotemporal diffusion with metric depth for 4d driving scene generation}},
  author={Guo, Jiazhe and Ding, Yikang and Chen, Xiwu and Chen, Shuo and Li, Bohan and Zou, Yingshuang and Lyu, Xiaoyang and Tan, Feiyang and Qi, Xiaojuan and Li, Zhiheng and others},
  journal={arXiv preprint arXiv:2503.15208},
  year={2025}
}

@article{ren2025cosmos,
  title={{Cosmos-Drive-Dreams: Scalable Synthetic Driving Data Generation with World Foundation Models}},
  author={Ren, Xuanchi and Lu, Yifan and Cao, Tianshi and Gao, Ruiyuan and Huang, Shengyu and Sabour, Amirmojtaba and Shen, Tianchang and Pfaff, Tobias and Wu, Jay Zhangjie and Chen, Runjian and others},
  journal={arXiv preprint arXiv:2506.09042},
  year={2025}
}

@inproceedings{gao2025magicdrive,
  title={{MagicDrive-V2: High-resolution long video generation for autonomous driving with adaptive control}},
  author={Gao, Ruiyuan and Chen, Kai and Xiao, Bo and Hong, Lanqing and Li, Zhenguo and Xu, Qiang},
  booktitle={Proceedings of the IEEE/CVF International Conference on Computer Vision},
  pages={28135--28144},
  year={2025}
}

@article{wan2025wan,
  title={{Wan: Open and advanced large-scale video generative models}},
  author={Wan, Team and Wang, Ang and Ai, Baole and Wen, Bin and Mao, Chaojie and Xie, Chen-Wei and Chen, Di and Yu, Feiwu and Zhao, Haiming and Yang, Jianxiao and others},
  journal={arXiv preprint arXiv:2503.20314},
  year={2025}
}

@inproceedings{trellis,
  title={{Structured 3d latents for scalable and versatile 3d generation}},
  author={Xiang, Jianfeng and Lv, Zelong and Xu, Sicheng and Deng, Yu and Wang, Ruicheng and Zhang, Bowen and Chen, Dong and Tong, Xin and Yang, Jiaolong},
  booktitle={Proceedings of the IEEE/CVF conference on computer vision and pattern recognition},
  pages={21469--21480},
  year={2025}
}

@article{ltx,
  title={{Ltx-video: Realtime video latent diffusion}},
  author={HaCohen, Yoav and Chiprut, Nisan and Brazowski, Benny and Shalem, Daniel and Moshe, Dudu and Richardson, Eitan and Levin, Eran and Shiran, Guy and Zabari, Nir and Gordon, Ori and others},
  journal={arXiv preprint arXiv:2501.00103},
  year={2024}
}

@inproceedings{DiT,
  title={Scalable diffusion models with transformers},
  author={Peebles, William and Xie, Saining},
  booktitle={Proceedings of the IEEE/CVF international conference on computer vision},
  pages={4195--4205},
  year={2023}
}

@inproceedings{neuroncap,
  title={Neuroncap: Photorealistic closed-loop safety testing for autonomous driving},
  author={Ljungbergh, William and Tonderski, Adam and Johnander, Joakim and Caesar, Holger and {\AA}str{\"o}m, Kalle and Felsberg, Michael and Petersson, Christoffer},
  booktitle={European Conference on Computer Vision},
  pages={161--177},
  year={2024},
  organization={Springer}
}

@article{fan2024lightgaussian,
  title={Lightgaussian: Unbounded 3d gaussian compression with 15x reduction and 200+ fps},
  author={Fan, Zhiwen and Wang, Kevin and Wen, Kairun and Zhu, Zehao and Xu, Dejia and Wang, Zhangyang},
  journal={Advances in neural information processing systems},
  volume={37},
  pages={140138--140158},
  year={2024}
}

@article{gao2026rad,
  title={Rad: Training an end-to-end driving policy via large-scale 3dgs-based reinforcement learning},
  author={Gao, Hao and Chen, Shaoyu and Jiang, Bo and Liao, Bencheng and Shi, Yiang and Guo, Xiaoyang and Pu, Yuechuan and Li, Xiangyu and Liu, Wenyu and Zhang, Qian and others},
  journal={Advances in Neural Information Processing Systems},
  volume={38},
  pages={32551--32576},
  year={2026}
}

@inproceedings{ren2026fastgs,
  title={Fastgs: Training 3d gaussian splatting in 100 seconds},
  author={Ren, Shiwei and Wen, Tianci and Fang, Yongchun and Lu, Biao},
  booktitle={Proceedings of the IEEE/CVF Conference on Computer Vision and Pattern Recognition},
  pages={26094--26103},
  year={2026}
}

@inproceedings{cheng2024rethinking,
  title={Rethinking imitation-based planners for autonomous driving},
  author={Cheng, Jie and Chen, Yingbing and Mei, Xiaodong and Yang, Bowen and Li, Bo and Liu, Ming},
  booktitle={2024 IEEE International Conference on Robotics and Automation (ICRA)},
  pages={14123--14130},
  year={2024},
  organization={IEEE}
}

@inproceedings{yang2024unipad,
  title={Unipad: A universal pre-training paradigm for autonomous driving},
  author={Yang, Honghui and Zhang, Sha and Huang, Di and Wu, Xiaoyang and Zhu, Haoyi and He, Tong and Tang, Shixiang and Zhao, Hengshuang and Qiu, Qibo and Lin, Binbin and others},
  booktitle={Proceedings of the IEEE/CVF conference on computer vision and pattern recognition},
  pages={15238--15250},
  year={2024}
}

@article{zhou2025hugsim,
  title={Hugsim: A real-time, photo-realistic and closed-loop simulator for autonomous driving},
  author={Zhou, Hongyu and Lin, Longzhong and Wang, Jiabao and Lu, Yichong and Bai, Dongfeng and Liu, Bingbing and Wang, Yue and Geiger, Andreas and Liao, Yiyi},
  journal={IEEE Transactions on Pattern Analysis and Machine Intelligence},
  year={2025},
  publisher={IEEE}
}

@inproceedings{yang2025drivearena,
  title={Drivearena: A closed-loop generative simulation platform for autonomous driving},
  author={Yang, Xuemeng and Wen, Licheng and Wei, Tiantian and Ma, Yukai and Mei, Jianbiao and Li, Xin and Lei, Wenjie and Fu, Daocheng and Cai, Pinlong and Dou, Min and others},
  booktitle={Proceedings of the IEEE/CVF International Conference on Computer Vision},
  pages={26933--26943},
  year={2025}
}

@inproceedings{yan2025drivingsphere,
  title={Drivingsphere: Building a high-fidelity 4d world for closed-loop simulation},
  author={Yan, Tianyi and Wu, Dongming and Han, Wencheng and Jiang, Junpeng and Zhou, Xia and Zhan, Kun and Xu, Cheng-zhong and Shen, Jianbing},
  booktitle={Proceedings of the Computer Vision and Pattern Recognition Conference},
  pages={27531--27541},
  year={2025}
}

@article{yan2026rlgf,
  title={Rlgf: Reinforcement learning with geometric feedback for autonomous driving video generation},
  author={Yan, Tianyi and Han, Wencheng and Zhang, Xueyang and Zhan, Kun and Xu, Cheng-Zhong and Shen, Jianbing and others},
  journal={Advances in Neural Information Processing Systems},
  volume={38},
  pages={128659--128684},
  year={2026}
}

@inproceedings{tang2025omnigen,
  title={Omnigen: Unified multimodal sensor generation for autonomous driving},
  author={Tang, Tao and Ma, Enhui and Zhou, Xia and Wang, Letian and Yan, Tianyi and Zhang, Xueyang and Zhan, Kun and Jia, Peng and Lang, Xianpeng and Bian, Jia-Wang and others},
  booktitle={Proceedings of the 33rd ACM International Conference on Multimedia},
  pages={9365--9374},
  year={2025}
}

@inproceedings{xia2026drivelaw,
  title={Drivelaw: Unifying planning and video generation in a latent driving world},
  author={Xia, Tianze and Li, Yongkang and Zhou, Lijun and Yao, Jingfeng and Xiong, Kaixin and Sun, Haiyang and Wang, Bing and Ma, Kun and Chen, Guang and Ye, Hangjun and others},
  booktitle={Proceedings of the IEEE/CVF Conference on Computer Vision and Pattern Recognition},
  pages={39701--39712},
  year={2026}
}

@inproceedings{deng2026gaussiandwm,
  title={Gaussiandwm: 3d gaussian driving world model for unified scene understanding and multi-modal generation},
  author={Deng, Tianchen and Chen, Xuefeng and Chen, Yi and Chen, Qu and Xu, Yuyao and Yang, Lijin and Xu, Le and Zhang, Yu and Zhang, Bo and Huang, Wuxiong and others},
  booktitle={Proceedings of the IEEE/CVF Conference on Computer Vision and Pattern Recognition},
  pages={10656--10667},
  year={2026}
}

@inproceedings{liang2026worldlens,
  title={WorldLens: Full-spectrum evaluations of driving world models in real world},
  author={Liang, Ao and Kong, Lingdong and Yan, Tianyi and Liu, Hongsi and Yang, Yu and Huang, Ziqi and Yin, Wei and Zuo, Jialong and Hu, Yixuan and Zhu, Dekai and others},
  booktitle={Proceedings of the IEEE/CVF Conference on Computer Vision and Pattern Recognition},
  pages={36385--36399},
  year={2026}
}

@article{yang2026x,
  title={X-scene: Large-scale driving scene generation with high fidelity and flexible controllability},
  author={Yang, Yu and Liang, Alan and Mei, Jianbiao and Ma, Yukai and Liu, Yong and Lee, Gim Hee},
  journal={Advances in Neural Information Processing Systems},
  volume={38},
  pages={104415--104451},
  year={2026}
}

@article{Huang2024S3GaussianSS,
  title={S3Gaussian: Self-Supervised Street Gaussians for Autonomous Driving},
  author={Nan Huang and Xiaobao Wei and Wenzhao Zheng and Pengju An and Ming Lu and Wei Zhan and Masayoshi Tomizuka and Kurt Keutzer and Shanghang Zhang},
  journal={ArXiv},
  year={2024},
  volume={abs/2405.20323}
}

@InProceedings{Zhou_2024_CVPR,
    author    = {Zhou, Hongyu and Shao, Jiahao and Xu, Lu and Bai, Dongfeng and Qiu, Weichao and Liu, Bingbing and Wang, Yue and Geiger, Andreas and Liao, Yiyi},
    title     = {HUGS: Holistic Urban 3D Scene Understanding via Gaussian Splatting},
    booktitle = {Proceedings of the IEEE/CVF Conference on Computer Vision and Pattern Recognition (CVPR)},
    month     = {June},
    year      = {2024},
    pages     = {21336-21345}
}

@InProceedings{Tonderski_2024_CVPR,
    author    = {Tonderski, Adam and Lindstr\"om, Carl and Hess, Georg and Ljungbergh, William and Svensson, Lennart and Petersson, Christoffer},
    title     = {NeuRAD: Neural Rendering for Autonomous Driving},
    booktitle = {Proceedings of the IEEE/CVF Conference on Computer Vision and Pattern Recognition (CVPR)},
    month     = {June},
    year      = {2024},
    pages     = {14895-14904}
}

@inproceedings{NEURIPS2024_46fd4317,
 author = {Fan, Zhiwen and Zhang, Jian and Cong, Wenyan and Wang, Peihao and Li, Renjie and Wen, Kairun and Zhou, Shijie and Kadambi, Achuta and Wang, Zhangyang and Xu, Danfei and Ivanovic, Boris and Pavone, Marco and Wang, Yue},
 booktitle = {Advances in Neural Information Processing Systems},
 doi = {10.52202/079017-1271},
 editor = {A. Globerson and L. Mackey and D. Belgrave and A. Fan and U. Paquet and J. Tomczak and C. Zhang},
 pages = {40212--40229},
 publisher = {Curran Associates, Inc.},
 title = {Large Spatial Model: End-to-end Unposed Images to Semantic 3D},
 volume = {37},
 year = {2024}
}

@inproceedings{ICLR2025_7dee643a,
 author = {Yang, Jiawei and Huang, Jiahui and Ivanovic, Boris and Chen, Yuxiao and Wang, Yan and Li, Boyi and You, Yurong and Sharma, Apoorva and Igl, Maximilian and Karkus, Peter and Xu, Danfei and Wang, Yue and Pavone, Marco},
 booktitle = {International Conference on Learning Representations},
 editor = {Y. Yue and A. Garg and N. Peng and F. Sha and R. Yu},
 pages = {50446--50465},
 title = {STORM: Spatio-TempOral Reconstruction Model For Large-Scale Outdoor Scenes},
 volume = {2025},
 year = {2025}
}

@InProceedings{Wen_2024_CVPR,
    author    = {Wen, Yuqing and Zhao, Yucheng and Liu, Yingfei and Jia, Fan and Wang, Yanhui and Luo, Chong and Zhang, Chi and Wang, Tiancai and Sun, Xiaoyan and Zhang, Xiangyu},
    title     = {Panacea: Panoramic and Controllable Video Generation for Autonomous Driving},
    booktitle = {Proceedings of the IEEE/CVF Conference on Computer Vision and Pattern Recognition (CVPR)},
    month     = {June},
    year      = {2024},
    pages     = {6902-6912}
}

@ARTICLE{11314796,
  author={Zhou, Hongyu and Lin, Longzhong and Wang, Jiabao and Lu, Yichong and Bai, Dongfeng and Liu, Bingbing and Wang, Yue and Geiger, Andreas and Liao, Yiyi},
  journal={IEEE Transactions on Pattern Analysis and Machine Intelligence}, 
  title={HUGSIM: A Real-Time, Photo-Realistic and Closed-Loop Simulator for Autonomous Driving}, 
  year={2026},
  volume={48},
  number={4},
  pages={4673-4691},
  doi={10.1109/TPAMI.2025.3647952}}

@InProceedings{Yang_2024_CVPR,
    author    = {Yang, Honghui and Zhang, Sha and Huang, Di and Wu, Xiaoyang and Zhu, Haoyi and He, Tong and Tang, Shixiang and Zhao, Hengshuang and Qiu, Qibo and Lin, Binbin and He, Xiaofei and Ouyang, Wanli},
    title     = {UniPAD: A Universal Pre-training Paradigm for Autonomous Driving},
    booktitle = {Proceedings of the IEEE/CVF Conference on Computer Vision and Pattern Recognition (CVPR)},
    month     = {June},
    year      = {2024},
    pages     = {15238-15250}
}

@inproceedings{NEURIPS2023_0c26a501,
 author = {Li, Quanyi and Peng, Zhenghao (Mark) and Feng, Lan and Liu, Zhizheng and Duan, Chenda and Mo, Wenjie and Zhou, Bolei},
 booktitle = {Advances in Neural Information Processing Systems},
 editor = {A. Oh and T. Naumann and A. Globerson and K. Saenko and M. Hardt and S. Levine},
 pages = {3894--3920},
 publisher = {Curran Associates, Inc.},
 title = {ScenarioNet: Open-Source Platform for Large-Scale Traffic Scenario Simulation and Modeling},
 volume = {36},
 year = {2023}
}

@ARTICLE{10643284,
  author={Jiang, Zhou and Zhu, Zhenxin and Li, Pengfei and Gao, Huan-ang and Yuan, Tianyuan and Shi, Yongliang and Zhao, Hang and Zhao, Hao},
  journal={IEEE Robotics and Automation Letters}, 
  title={P-MapNet: Far-Seeing Map Generator Enhanced by Both SDMap and HDMap Priors}, 
  year={2024},
  volume={9},
  number={10},
  pages={8539-8546},
  doi={10.1109/LRA.2024.3447450}}

@InProceedings{Lu_2025_ICCV,
    author    = {Lu, Yifan and Ren, Xuanchi and Yang, Jiawei and Shen, Tianchang and Wu, Zhangjie and Gao, Jun and Wang, Yue and Chen, Siheng and Chen, Mike and Fidler, Sanja and Huang, Jiahui},
    title     = {InfiniCube: Unbounded and Controllable Dynamic 3D Driving Scene Generation with World-Guided Video Models},
    booktitle = {Proceedings of the IEEE/CVF International Conference on Computer Vision (ICCV)},
    month     = {October},
    year      = {2025},
    pages     = {27272-27283}
}

@ARTICLE{10517470,
  author={Pan, Jingyi and Wang, Zipeng and Wang, Lin},
  journal={IEEE Robotics and Automation Letters}, 
  title={Co-Occ: Coupling Explicit Feature Fusion With Volume Rendering Regularization for Multi-Modal 3D Semantic Occupancy Prediction}, 
  year={2024},
  volume={9},
  number={6},
  pages={5687-5694},
  doi={10.1109/LRA.2024.3396092}}

@InProceedings{Song_2025_ICCV,
    author    = {Song, Rui and Liang, Chenwei and Xia, Yan and Zimmer, Walter and Cao, Hu and Caesar, Holger and Festag, Andreas and Knoll, Alois},
    title     = {CoDa-4DGS: Dynamic Gaussian Splatting with Context and Deformation Awareness for Autonomous Driving},
    booktitle = {Proceedings of the IEEE/CVF International Conference on Computer Vision (ICCV)},
    month     = {October},
    year      = {2025},
    pages     = {28031-28041}
}

@InProceedings{Kung_2025_ICCV,
    author    = {Kung, Pou-Chun and Harisha, Skanda and Vasudevan, Ram and Eid, Aline and Skinner, Katherine A.},
    title     = {RadarSplat: Radar Gaussian Splatting for High-Fidelity Data Synthesis and 3D Reconstruction of Autonomous Driving Scenes},
    booktitle = {Proceedings of the IEEE/CVF International Conference on Computer Vision (ICCV)},
    month     = {October},
    year      = {2025},
    pages     = {27596-27606}
}

@InProceedings{Lindstrom_2024_CVPR,
    author    = {Lindstr\"om, Carl and Hess, Georg and Lilja, Adam and Fatemi, Maryam and Hammarstrand, Lars and Petersson, Christoffer and Svensson, Lennart},
    title     = {Are NeRFs Ready for Autonomous Driving? Towards Closing the Real-to-simulation Gap},
    booktitle = {Proceedings of the IEEE/CVF Conference on Computer Vision and Pattern Recognition (CVPR) Workshops},
    month     = {June},
    year      = {2024},
    pages     = {4461-4471}
}

@InProceedings{Xu_2025_ICCV,
    author    = {Xu, Jiawei and Deng, Kai and Fan, Zexin and Wang, Shenlong and Xie, Jin and Yang, Jian},
    title     = {AD-GS: Object-Aware B-Spline Gaussian Splatting for Self-Supervised Autonomous Driving},
    booktitle = {Proceedings of the IEEE/CVF International Conference on Computer Vision (ICCV)},
    month     = {October},
    year      = {2025},
    pages     = {24770-24779}
}

@inproceedings{10.1145/3664647.3681482,
author = {Tao, Tang and Gao, Longfei and Wang, Guangrun and Lao, Yixing and Chen, Peng and Zhao, Hengshuang and Hao, Dayang and Liang, Xiaodan and Salzmann, Mathieu and Yu, Kaicheng},
title = {LiDAR-NeRF: Novel LiDAR View Synthesis via Neural Radiance Fields},
year = {2024},
isbn = {9798400706868},
publisher = {Association for Computing Machinery},
address = {New York, NY, USA},
url = {https://doi.org/10.1145/3664647.3681482},
doi = {10.1145/3664647.3681482},
booktitle = {Proceedings of the 32nd ACM International Conference on Multimedia},
pages = {390–398},
numpages = {9},
location = {Melbourne VIC, Australia},
series = {MM '24}
}

@INPROCEEDINGS{11127564,
  author={Khan, Mustafa and Fazlali, Hamidreza and Sharma, Dhruv and Cao, Tongtong and Bai, Dongfeng and Ren, Yuan and Liu, Bingbing},
  booktitle={2025 IEEE International Conference on Robotics and Automation (ICRA)}, 
  title={AutoSplat: Constrained Gaussian Splatting for Autonomous Driving Scene Reconstruction}, 
  year={2025},
  volume={},
  number={},
  pages={8315-8321},
  doi={10.1109/ICRA55743.2025.11127564}}

@INPROCEEDINGS{11127463,
  author={Mao, Jiageng and Li, Boyi and Ivanovic, Boris and Chen, Yuxiao and Wang, Yan and You, Yurong and Xiao, Chaowei and Xu, Danfei and Pavone, Marco and Wang, Yue},
  booktitle={2025 IEEE International Conference on Robotics and Automation (ICRA)}, 
  title={DreamDrive: Generative 4D Scene Modeling from Street View Images}, 
  year={2025},
  volume={},
  number={},
  pages={367-374},
  doi={10.1109/ICRA55743.2025.11127463}}

@InProceedings{Xu_2025_CVPR,
    author    = {Xu, Zhenhua and Bai, Yan and Zhang, Yujia and Li, Zhuoling and Xia, Fei and Wong, Kwan-Yee K. and Wang, Jianqiang and Zhao, Hengshuang},
    title     = {DriveGPT4-V2: Harnessing Large Language Model Capabilities for Enhanced Closed-Loop Autonomous Driving},
    booktitle = {Proceedings of the IEEE/CVF Conference on Computer Vision and Pattern Recognition (CVPR)},
    month     = {June},
    year      = {2025},
    pages     = {17261-17270}
}

@InProceedings{10.1007/978-3-031-72943-0_15,
author="Sima, Chonghao
and Renz, Katrin
and Chitta, Kashyap
and Chen, Li
and Zhang, Hanxue
and Xie, Chengen
and Bei{\ss}wenger, Jens
and Luo, Ping
and Geiger, Andreas
and Li, Hongyang",
editor="Leonardis, Ale{\v{s}}
and Ricci, Elisa
and Roth, Stefan
and Russakovsky, Olga
and Sattler, Torsten
and Varol, G{\"u}l",
title="DriveLM: Driving with Graph Visual Question Answering",
booktitle="Computer Vision -- ECCV 2024",
year="2025",
publisher="Springer Nature Switzerland",
address="Cham",
pages="256--274",
isbn="978-3-031-72943-0"
}

@inproceedings{NEURIPS2025_6e4f0c8c,
 author = {Chi, Haohan and Gao, Huan-ang and Liu, Ziming and Liu, Jianing and Liu, Chenyu and Li, Jinwei and Yang, Kaisen and Yu, Yangcheng and Wang, Zeda and Li, Wenyi and Wang, Leichen and HU, Xingtao and SUN, HAO and Zhao, Hang and Zhao, Hao},
 booktitle = {Advances in Neural Information Processing Systems},
 editor = {D. Belgrave and C. Zhang and H. Lin and R. Pascanu and P. Koniusz and M. Ghassemi and N. Chen},
 pages = {},
 publisher = {Curran Associates, Inc.},
 title = {Impromptu VLA: Open Weights and Open Data for Driving Vision-Language-Action Models},
 volume = {38},
 year = {2025}
}

@article{HUANG2025105321,
title = {VLM-RL: A unified vision language models and reinforcement learning framework for safe autonomous driving},
journal = {Transportation Research Part C: Emerging Technologies},
volume = {180},
pages = {105321},
year = {2025},
issn = {0968-090X},
doi = {https://doi.org/10.1016/j.trc.2025.105321},
url = {https://www.sciencedirect.com/science/article/pii/S0968090X25003250},
author = {Zilin Huang and Zihao Sheng and Yansong Qu and Junwei You and Sikai Chen}
}

@ARTICLE{11394788,
  author={Zheng, Yupeng and Xing, Zebin and Zhang, Qichao and Jin, Bu and Li, Pengfei and Zheng, Yuhang and Xia, Zhongpu and Chen, Yaran and Zhao, Dongbin},
  journal={IEEE Transactions on Cognitive and Developmental Systems}, 
  title={PlanAgent: A Multi-modal Large Language Agent for Closed-loop Vehicle Motion Planning}, 
  year={2026},
  volume={},
  number={},
  pages={1-14},
  doi={10.1109/TCDS.2026.3664120}}

@INPROCEEDINGS{10611018,
  author={Chen, Long and Sinavski, Oleg and Hünermann, Jan and Karnsund, Alice and Willmott, Andrew James and Birch, Danny and Maund, Daniel and Shotton, Jamie},
  booktitle={2024 IEEE International Conference on Robotics and Automation (ICRA)}, 
  title={Driving with LLMs: Fusing Object-Level Vector Modality for Explainable Autonomous Driving}, 
  year={2024},
  volume={},
  number={},
  pages={14093-14100},
  doi={10.1109/ICRA57147.2024.10611018}}

@InProceedings{Zheng_2025_ICCV,
    author    = {Zheng, Yupeng and Yang, Pengxuan and Xing, Zebin and Zhang, Qichao and Zheng, Yuhang and Gao, Yinfeng and Li, Pengfei and Zhang, Teng and Xia, Zhongpu and Jia, Peng and Lang, XianPeng and Zhao, Dongbin},
    title     = {World4Drive: End-to-End Autonomous Driving via Intention-aware Physical Latent World Model},
    booktitle = {Proceedings of the IEEE/CVF International Conference on Computer Vision (ICCV)},
    month     = {October},
    year      = {2025},
    pages     = {28632-28642}
}

@InProceedings{10.1007/978-3-031-72995-9_23,
author="Ma, Yingzi
and Cao, Yulong
and Sun, Jiachen
and Pavone, Marco
and Xiao, Chaowei",
editor="Leonardis, Ale{\v{s}}
and Ricci, Elisa
and Roth, Stefan
and Russakovsky, Olga
and Sattler, Torsten
and Varol, G{\"u}l",
title="Dolphins: Multimodal Language Model for Driving",
booktitle="Computer Vision -- ECCV 2024",
year="2025",
publisher="Springer Nature Switzerland",
address="Cham",
pages="403--420",
isbn="978-3-031-72995-9"
}

@misc{xu2025vlmadendtoendautonomousdriving,
      title={VLM-AD: End-to-End Autonomous Driving through Vision-Language Model Supervision}, 
      author={Yi Xu and Yuxin Hu and Zaiwei Zhang and Gregory P. Meyer and Siva Karthik Mustikovela and Siddhartha Srinivasa and Eric M. Wolff and Xin Huang},
      year={2025},
      eprint={2412.14446},
      archivePrefix={arXiv},
      primaryClass={cs.CV},
      url={https://arxiv.org/abs/2412.14446}, 
}

@InProceedings{Jiang_2025_ICCV,
    author    = {Jiang, Sicong and Huang, Zilin and Qian, Kangan and Luo, Ziang and Zhu, Tianze and Zhong, Yang and Tang, Yihong and Kong, Menglin and Wang, Yunlong and Jiao, Siwen and Ye, Hao and Sheng, Zihao and Zhao, Xin and Wen, Tuopu and Fu, Zheng and Chen, Sikai and Jiang, Kun and Yang, Diange and Choi, Seongjin and Sun, Lijun},
    title     = {A Survey on Vision-Language-Action Models for Autonomous Driving},
    booktitle = {Proceedings of the IEEE/CVF International Conference on Computer Vision (ICCV) Workshops},
    month     = {October},
    year      = {2025},
    pages     = {4583-4595}
}

@InProceedings{Liu_2026_CVPR,
    author    = {Liu, Lin and Jia, Caiyan and Yu, Guanyi and Song, Ziying and Li, Junqiao and Jia, Feiyang and Wu, Peiliang and Hao, Xiaoshuai and Luo, Yadan},
    title     = {GuideFlow: Constraint-Guided Flow Matching for Planning in End-to-End Autonomous Driving},
    booktitle = {Proceedings of the IEEE/CVF Conference on Computer Vision and Pattern Recognition (CVPR)},
    month     = {June},
    year      = {2026},
    pages     = {3719-3728}
}

@ARTICLE{11457610,
  author={Wozniak, Maciej and Liu, Lianhang and Cai, Yixi and Jensfelt, Patric},
  journal={IEEE Robotics and Automation Letters}, 
  title={ PRIX: Learning to Plan From Raw Pixels for End-to-End Autonomous Driving}, 
  year={2026},
  volume={11},
  number={5},
  pages={6400-6407},
  doi={10.1109/LRA.2026.3678836}}

@misc{li2025drivevlaw0worldmodelsamplify,
      title={DriveVLA-W0: World Models Amplify Data Scaling Law in Autonomous Driving}, 
      author={Yingyan Li and Shuyao Shang and Weisong Liu and Bing Zhan and Haochen Wang and Yuqi Wang and Yuntao Chen and Xiaoman Wang and Yasong An and Chufeng Tang and Lu Hou and Lue Fan and Zhaoxiang Zhang},
      year={2025},
      eprint={2510.12796},
      archivePrefix={arXiv},
      primaryClass={cs.CV},
      url={https://arxiv.org/abs/2510.12796}, 
}

@InProceedings{Rawal_2026_CVPR,
    author    = {Rawal, Ishaan and Gupta, Shubh and Hu, Yihan and Zhan, Wei},
    title     = {NoRD: A Data-Efficient Vision-Language-Action Model that Drives without Reasoning},
    booktitle = {Proceedings of the IEEE/CVF Conference on Computer Vision and Pattern Recognition (CVPR)},
    month     = {June},
    year      = {2026},
    pages     = {10965-10975}
}

\end{document}